\documentclass[10pt,journal,cspaper,compsoc]{IEEEtran}
\ifCLASSOPTIONcompsoc
\else
\fi
\usepackage{url}
\usepackage{algorithm}
\usepackage{algorithmic}
\usepackage{booktabs}
\usepackage{tabularx}

\usepackage{multirow}

\ifCLASSINFOpdf
   \usepackage[pdftex]{graphicx}

\else
  \usepackage[dvips]{graphicx}
\fi

\usepackage[cmex10]{amsmath}

\usepackage{algorithmic}

\usepackage{array}

\usepackage{mdwmath}
\usepackage{mdwtab}

\usepackage{eqparbox}

\usepackage[tight,footnotesize]{subfigure}

 \let\MYorigsubfigure\subfigure
 \renewcommand{\subfigure}[2][\relax]{\MYorigsubfigure[]{#2}}

\usepackage{epsfig}
\usepackage{graphicx}
\usepackage{amsmath}
\usepackage{amssymb}
\usepackage{extarrows }
\usepackage{subfigure}
\usepackage{overpic}

\usepackage{caption}
\begin{document}
\title{Visual Saliency Based on Scale-Space Analysis in the Frequency Domain}

\author{Jian~Li,~\IEEEmembership{Student Member,~IEEE,}
        Martin~D. Levine,~\IEEEmembership{Fellow,~IEEE,}\\
        Xiangjing~An,~\IEEEmembership{Member,~IEEE,}
        Xin~Xu~\IEEEmembership{Member,~IEEE}
        and~Hangen~He~\IEEEmembership{}
\IEEEcompsocitemizethanks{\IEEEcompsocthanksitem Jian Li, Xiangjing An, Xin Xu and Hangen He are with the Institute of Automation, National University of Defense Technology, Changsha,
PR China, 410073. E-mail: lijian@nudt.edu.cn.
\IEEEcompsocthanksitem Martin D. Levine is with the Centre for Intelligent Machines, McGill University,
3480 University Street, Montreal, Quebec, Canada H3A 2A7.
E-mail: levine@cim.mcgill.ca
.}
\thanks{}}

\markboth{Journal of \LaTeX\ Class Files,~Vol.~6, No.~1, JANUARY~2007}%
{Shell \MakeLowercase{\textit{et al.}}: Bare Demo of IEEEtran.cls for Computer Society Journals}

\IEEEcompsoctitleabstractindextext{%
\begin{abstract}

We address the issue of visual saliency from three perspectives. First, we consider saliency detection as a frequency domain analysis problem. Second, we achieve this by employing the concept of {\it non-saliency}. Third, we simultaneously consider the detection of salient regions of different size. The paper proposes a new bottom-up paradigm for detecting visual saliency, characterized by a scale-space analysis of the amplitude spectrum of natural images. We show that the convolution of the {\it image amplitude spectrum} with a low-pass Gaussian kernel of an appropriate scale is equivalent to such an image saliency detector. The saliency map is obtained by reconstructing the 2-D signal using the original phase and the amplitude spectrum, filtered at a scale selected by minimizing saliency map entropy. A Hypercomplex Fourier Transform performs the analysis in the frequency domain. Using available databases, we demonstrate experimentally that the proposed model can predict human fixation data. We also introduce a new image database and use it to show that the saliency detector can highlight both small and large salient regions, as well as inhibit repeated distractors in cluttered images. In addition, we show that it is able to predict salient regions on which people focus their attention.

\end{abstract}

\begin{IEEEkeywords}
Visual attention, saliency, Hypercomplex Fourier Transform, eye-tracking, scale space analysis.
\end{IEEEkeywords}}

\maketitle
\IEEEdisplaynotcompsoctitleabstractindextext

\IEEEpeerreviewmaketitle

\section{Introduction}
\label{Intro}

\IEEEPARstart{V}{isual} attention facilitates our ability to rapidly locate the most important information in a scene \cite{ Yarbus1967eye,neisser1967cognitive}. Such image regions are said to be salient since it is assumed that they attract greater attention by the visual system than other parts of the image. These sali-ent regions are expected to possess distinctive features when compared with others in the image. The study of saliency detection may reveal the attentional mechanisms of biological visual systems, as well as model their fixation selection behavior. On the other hand, as a component of low-level artificial vision processing, it facilitates subsequent processing such as object detection or recognition by reducing computational cost, which is a key consideration in real-time applications. For object detection, this would always be more efficient than dense sampling, provided one could ensure the accuracy of the attentional mechanism.

Visual saliency detection has received extensive attention by both psychologists and computer vision researchers \cite{Itti_etal98pami, Itti_Koch01nrn, tsotsos2008roles, NIPS2005_81, saliencypoggio, NIPS2006_897, le2006coherent,kienzle2007nonparametric, mahadevan2009spatiotemporal, NIPS2007_874, NIPS2007_1074, tsotsos2011computational, NIPS2008_0142, ittisurprise}, and many models have been proposed based on different assumptions. Generally speaking, there are two different processes that influence visual saliency; one is top-down and depends on the task at hand and the other is bottom-up, which is driven by the input image. The focus of the paper is bottom-up saliency for selecting attentional regions.

Many bottom-up computational models that simulate primate perceptual abilities have appeared in the literature. For example, in \cite{Itti_etal98pami, Itti_Koch01nrn, itti2000saliency} a center-surround mechanism is used to define saliency across scales, which is inspired by the putative neural mechanism. It has also been hypothesized that some visual inputs are intrinsically salient in certain background contexts and that these are actually task-independent \cite{Itti_etal98pami, Itti_Koch01nrn}. This model has established itself as the exemplar for saliency detection and consistently used for comparison in the literature. Similarly, there are also several proposed models which use other types of local information in different ways. In \cite{kadir2001saliency}, saliency is defined as the local complexity. Gao {\it et al.} \cite{gao2009discriminant,gao2007bottom,gao2008plausibility} proposed a bottom-up saliency model by using Kullback-Leibler (KL) divergence to measure the difference between a location and its surrounding area. In \cite{NIPS2005_81}, a model of overt attention with selection based on self-information is proposed, where the patches of an image are decomposed into a set of pre-learned bases and kernel density estimation is used to approximate the self-information.

Several models have been suggested to compute saliency using global information. In \cite{LCAV-CONF-2009-012}, the authors first transform the input color image into the $Lab$ color space and then define the saliency at each location as the difference between the $Lab$ pixel value and the mean $Lab$ value of the entire image. Harel {\it et al.} \cite{NIPS2006_897} proposed a graph-based solution that uses local computation to obtain a saliency map, which is everywhere dependent on global information. In \cite{avraham2009esaliency}, a saliency model called "extended saliency" was proposed, in which the "global exceptions" concept is used to replace the traditional preference for local contrast. Recently, a simple and fast algorithm, called the spectrum residual (SR), was proposed based on the Fourier Transform \cite{hou2007saliency}. The paper argues that the {\it spectrum residual} corresponds to image saliency. Following this, the Phase spectrum of the Fourier Transform (PFT) was introduced, which achieved nearly the same performance as the SR \cite{guo2008spatio}. Based on PFT, PQFT \cite{guo2008spatio} was also proposed by combining more features and using the quaternion Fourier Transform.

In this paper, we address the issue from three perspectives. Inspired by \cite{hou2007saliency}, we first consider saliency detection as a frequency domain problem. Unlike recent approaches which model saliency as a local phenomenon, we propose a new frequency domain para-digm, which permits the full use of global information. Second, instead of modeling saliency in an image, we define the concept of {\it non-saliency} using global information. Although in this paper we are solely concerned with determining saliency  computationally, it is also interesting to consider the biological point of view. Research has suggested that objects viewed by the human visual system are thought to compete with each other to selectively focus our attention on a subset \cite{beck2005stimulus, yantis2005visual}. Objects that appear in the visual field will influence how they are viewed by suppressing each other. Consequently, many are inhibited, while those that are not, will ultimately predominate in the visual cortex to provide a focus of attention. In this paper, we model these inhibited regions as non-saliency. Compared with salient regions, which are very distinctive in the image, non-saliency can usually be modeled by common or uniform regions. These are then  suppressed,  thereby permitting salient objects to literally pop out. In this paper, non-saliency is modeled in the frequency domain. Third, we also address another issue, that of detecting salient regions of different sizes.
To date there is no consistent definition of saliency in the literature. The models of saliency are diverse. In several models, saliency detection mimics the fixation selection mechanism and tends to find small distinct regions or points, for example, SR \cite{hou2007saliency}, PFT \cite{guo2008spatio}, PQFT \cite{guo2010multiresolution} and AIM \cite{NIPS2005_81}. However, these may fail when detecting large saliency regions. Other papers tend to find large salient regions \cite{LCAV-CONF-2009-012, liu2010learning, goferman2010context, chengsaliency,yusaliencyregions, khan2009top}. Recently, scale-aware saliency \cite{Jacobson2010scale-aware} has been introduced to alleviate the problem of fixed scale in the spatial domain. We consider both small salient points as well as salient regions. For convenience, we will refer to both of these as salient regions, but of different size. We will show that the size of saliency regions is related to a scale parameter in the frequency domain.

We propose a new framework for saliency detection, which ostensibly, at first sight, seems to be similar to the {\it Convolution Theorem} but in fact we will show that it is not. We will demonstrate that the convolution of the amplitude spectrum with a Gaussian kernel of an appropriate scale is equivalent to a saliency detector. The proposed framework has the ability to both highlight small and large salient regions and to inhibit repeated distractors in cluttered images.

The contribution of this paper is threefold: 1) A frequency domain paradigm for saliency detection is proposed; 2) The detection of both small and large salient regions is treated as a whole in the proposed model; 3) We show that SR, PFT and the frequency-tuned model\cite{LCAV-CONF-2009-012} are, to some extent, special cases of the proposed model.

The paper is organized as follows: Section 2 is the description and review of related work. In section 3, we present the theoretical background of the proposed framework, called Spectrum Scale Space analysis (SSS). We present the saliency model (HFT) based on the Hypercomplex Fourier Transform in section 4. In section 5, we discuss experimental results. Concluding remarks and possible extensions are discussed in section 6.

\section{Related Work}
\label{relatedworks}

Recently, a simple and fast algorithm, called the {\it Spectrum Residual} (SR) was proposed in \cite{hou2007saliency}. This paper argues that the spectrum residual corresponds to image saliency. Thus given an image $f(x,y)$, it was first transformed into the frequency domain: $f(x,y)\xlongrightarrow{\mathcal F}{\mathcal F}(f)(u,v) $.
The amplitude ${\mathcal A}(u,v)=|{\mathcal F}(f)|$ and phase ${\mathcal P}(u,v) =angle({\mathcal F}(f))$ spectra are calculated, and then the log amplitude spectrum is obtained: ${\mathcal L}(u,v)=log({\mathcal A}(u,v))$. Given these definitions, the spectrum residual was defined as:
\begin{equation}
{\mathcal R}(u,v)={\mathcal L}(u,v)-h_{n}\star{\mathcal L}(u,v),  \label {equ2}
\end {equation}
and the saliency map ${\mathcal S}(x,y)$ of the original image as:
\begin{equation}
{\mathcal S}(x,y)={\mathcal F}^{-1}[exp({\mathcal R}(u,v)+i\cdot{\mathcal P}(u,v))].  \label {equ3}
\end {equation}
In order to obtain a better visual display, the final saliency map was actually presented as\footnote{In this paper, $\vert\cdot\vert^2$ indicates computing the square of each element in the matrix.}:
\begin{equation}
{\mathcal S}(x,y)=g\star\vert{\mathcal F}^{-1}[exp({\mathcal R}(u,v)+i\cdot{\mathcal P}(u,v))]\vert^2,  \label {equ4}
\end {equation}
where ${\mathcal F}$ and ${\mathcal F}^{-1}$ denote the Fourier and inverse Fourier Transforms, respectively; $h_{n}$ and $g$ are low-pass filters;  $i$ is the imaginary function; ${\mathcal P}(u,v)$ denotes the phase spectrum of the image, {\it which is assumed to be preserved} when transforming back to the spatial domain. Equations (\ref{equ2}-\ref{equ4}) are from \cite{hou2007saliency}. The spectrum residual is the key idea of the SR, and the authors argued that it is this residual, combined with the original phase spectrum, that corresponds to the saliency in the image. However, in this paper:
1) We will show that the spectrum residual is of little significance;
2) For natural images, SR (or other similar models such as PFT \cite{guo2008spatio})  are, to some extent, equivalent to a gradient operator; and
3) Provide an explanation of why SR works in certain cases.

\begin{figure}
\begin{center}
   \includegraphics[width=.82\linewidth]{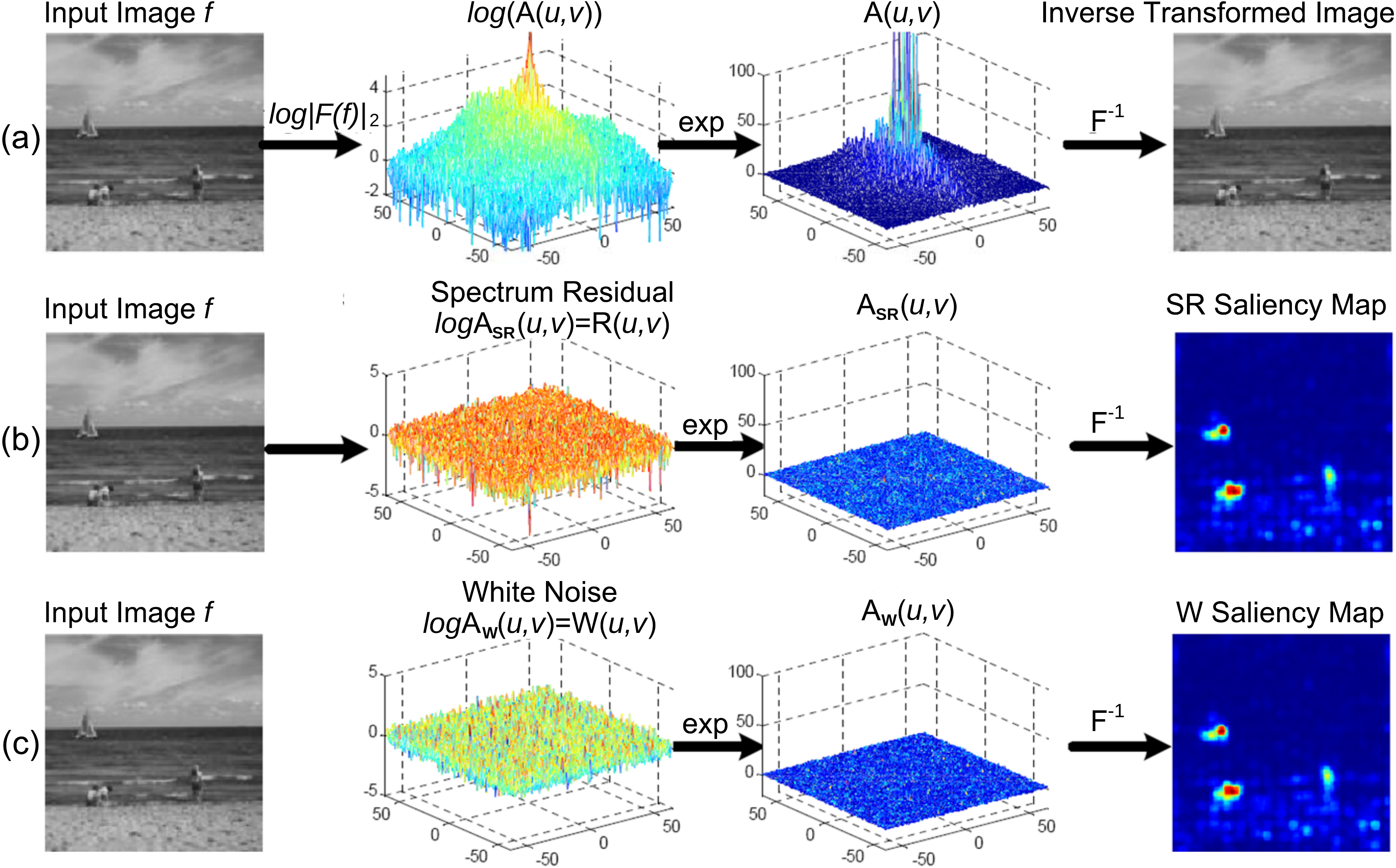}
 \end{center}
  \caption{Spectrum residual given by SR contains little information corresponding to image saliency. (a) Obviously, the original image is reproduced by performing the inverse FT using the original amplitude and phase spectrum. (b) In SR, it is argued that saliency map can be obtained by replacing the $log({\mathcal A}(u,v))$ by the Spectrum Residual ${\mathcal R}(u,v)$. (c) If we replace the log amplitude spectrum $log{\mathcal A}(u,v)$ by random white noise(sic), we can obtain nearly the same saliency map.}
  \label{fig:sr}
\end{figure}

For convenience, we rewrite the standard {\it inverse Fourier Transform} as follows:
\begin{eqnarray}
&f(x,y)&={\mathcal F}^{-1}[exp(log {\mathcal A}(u,v)+i\cdot{\mathcal P}(u,v))],  \label {equ5} \\
\Leftrightarrow&f(x,y)&={\mathcal F}^{-1}[{\mathcal A}(u,v)\cdot exp(i\cdot{\mathcal P}(u,v))],   \label {equ6}\\
\Leftrightarrow&f(x,y)&={\mathcal F}^{-1}[{\mathcal F}(f)(u,v)].   \label {equ7}
\end{eqnarray}
Thus we can rewrite (\ref{equ3}) as follows:
\begin{equation}
{\mathcal S}(x,y)={\mathcal F}^{-1}[exp({\mathcal R}(u,v)\cdot exp(i\cdot{\mathcal P}(u,v))],  \label {equ8}
\end {equation}
Defining $exp({\mathcal R}(u,v))$ as ${\mathcal A}_{SR}(u,v)$, (\ref{equ8}) is rewritten as:
\begin{equation}
{\mathcal S}(x,y)={\mathcal F}^{-1}[{\mathcal A}_{SR}(u,v)\cdot exp(i\cdot{\mathcal P}(u,v))].  \label {equ9}
\end {equation}
Comparing (\ref{equ6}) and (\ref{equ9}), we observe that if we replace the amplitude spectrum ${\mathcal A}(u,v)$ by the exponential of ${\mathcal R}(u,v)$, the saliency map is obtained\footnote {The phase spectra will no longer be plotted in the remaining figures in this paper, although, obviously they exist and are required for computing the transforms.}. (See the comparison in Fig. \ref{fig:sr}(a, b)).  This is the key idea of SR.

%

In order to illustrate that the spectrum residual is of little significance, we generate a 2D white noise signal ${\mathcal W}(u,v)$, which has the same average value and maximum as the spectrum residual ${\mathcal R}(u,v)$. We then use ${\mathcal W}(u,v)$ to replace the spectrum residual and perform the inverse Fourier Transform as follows:
\begin{equation}
{\mathcal S}(x,y)={\mathcal F}^{-1}[exp({\mathcal W}(u,v)\cdot exp(i\cdot{\mathcal P}(u,v))].  \label {equ10}
\end{equation}
Fig.\ref{fig:sr}(c) shows this process. If we define $exp({\mathcal W}(u,v))$ as ${\mathcal A}_{W}(u,v)$, (\ref{equ10}) can be rewritten as follows:
\begin{equation}
{\mathcal S}(x,y)={\mathcal F}^{-1}[{\mathcal A}_{W}(u,v)\cdot exp(i\cdot{\mathcal P}(u,v))].  \label {equ11}
\end {equation}
Surprisingly, we can obtain nearly the same saliency map when we use white noise to replace the spectrum residual. This result clearly shows that the spectrum residual in \cite{hou2007saliency} contains little information corresponding to saliency. Why is this the case? Comparing (\ref{equ9}) and (\ref{equ11}), we find that the amplitude spectra used to perform the inverse Fourier Transform are $A_{SR}(u,v)$ and ${\mathcal A}_{W}(u,v)$. As shown in the third columns of Fig.{\ref{fig:sr}}(b, c), both ${\mathcal A}_{SR}(u,v)$ and ${\mathcal A}_{W}(u,v)$ are nearly horizontal planes compared  (at the same scale) with ${\mathcal A}(u,v)$ shown in Fig.\ref{fig:sr}(a). That is to say, in both (\ref{equ9}) and (\ref{equ11}), the amplitude information is totally abandoned and only phase information plays a role.

Two questions arise:
(1) Why does SR yield a saliency map using only phase information?
(2) More important, is there any information corresponding to image saliency contained in the amplitude spectrum?
For the first question, our answer is that it only works for certain cases (detecting small salient regions in uncluttered scenes).

Also, consider \cite{guo2008spatio, guo2010multiresolution} where the authors propose a new saliency model called the Phase Fourier Transform (PFT). The saliency is computed using only phase information as follows:
\begin{eqnarray}
&{\mathcal S}(x,y)&={\mathcal F}^{-1}[exp(i\cdot{\mathcal P}(u,v))],  \label {equ12} \\
\Leftrightarrow&{\mathcal S}(x,y)&={\mathcal F}^{-1}[1(u,v)\cdot exp(i\cdot{\mathcal P}(u,v))].   \label {equ14}
\end{eqnarray}
We observe that in PFT, the amplitude spectrum is (implicitly) also replaced by a horizontal plane. Therefore, we can deduce that, for natural images, both SR and PFT will produce nearly the same saliency map.

\begin{figure}[h]
\begin{center}
\begin{overpic}[width=7cm]{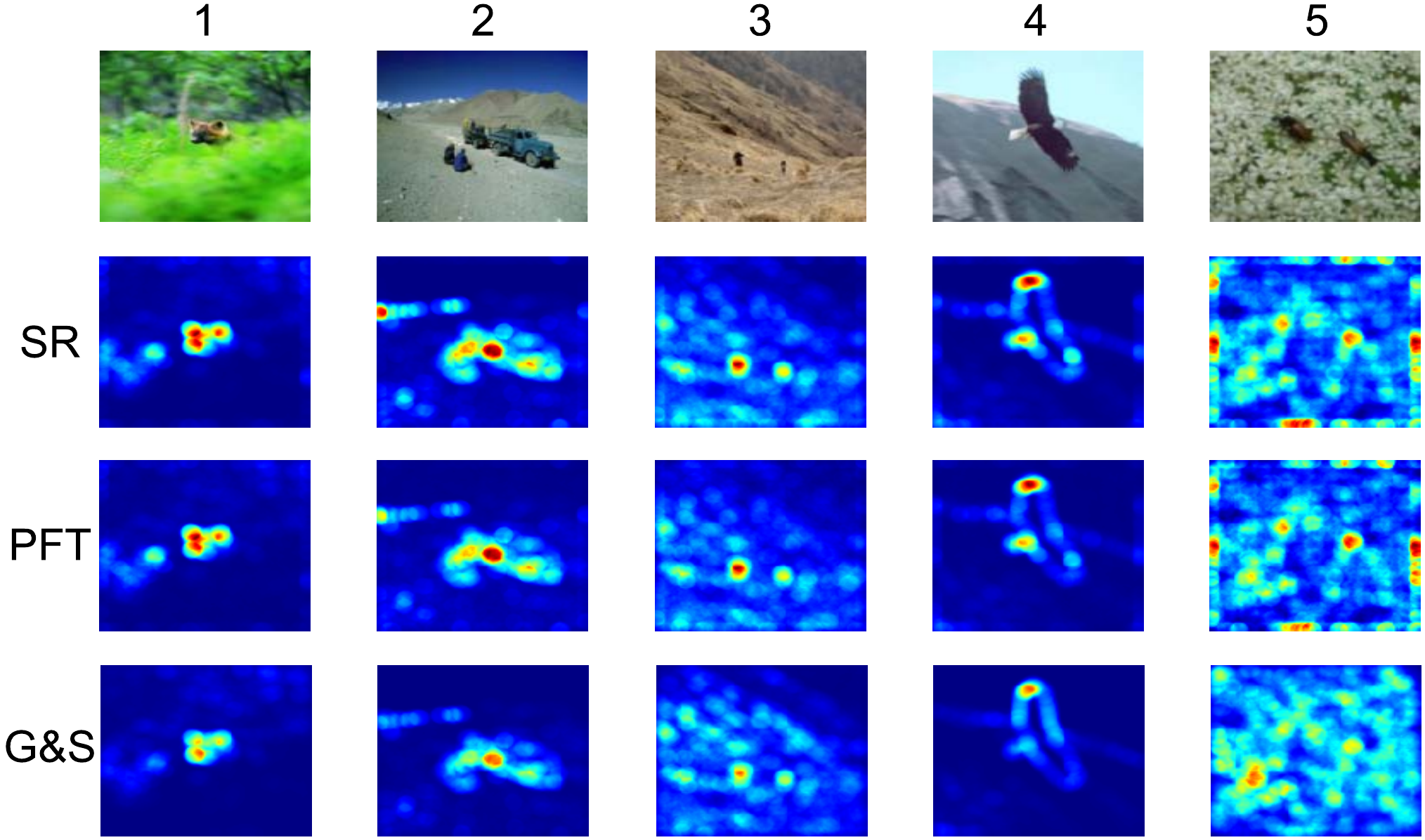}
\end{overpic}
\caption{G\&S achieves nearly the same performance as SR and PFT.}
\label{fig: SRvsGS}
\end{center}
\end{figure}

What does using the inverse Fourier Transform solely with phase information imply? We argue that for natural images, both SR and PFT are, to a certain degree, equivalent to a gradient operator combined with Gaussian post-processing (like the $g$ in (\ref{equ4})). This is because the amplitude spectrum of natural images always has higher values at low than at high frequencies \cite{ruderman1994statistics, srivastava2003advances}. Thus if the amplitude spectrum is replaced by a horizontal plane, all of the frequencies are being treated equally. That is to say, the lower frequencies are suppressed and the higher frequencies are enhanced. As is well-known, this implies a gradient enhancement operation. Based on the above discussion, we conclude that both SR and PFT will enhance the object boundaries and textured parts in an image. This indicates that they could work well only in detecting small salient regions where the center-surround contrast is very strong (see col. 1 and 3 of Fig. \ref{fig: SRvsGS}). However, they will have difficulty detecting large salient regions (col. 4) and those in a cluttered background  (col. 5). To illustrate this point, we use a simple gradient operation combined with Gaussian post-filtering (as given by Algorithm \ref{alg:GM} below) and obtain nearly the same performance as the other two methods, as shown in Fig. \ref{fig: SRvsGS}.

Why is the performance of these models inadequate? The reason is that the information contained in the amplitude spectrum has been abandoned.
\begin{algorithm}[htb]         
\caption{Procedure for computing the {\bf gradient and smoothing(G\&S)} }             
\label{alg:GM}                  
\begin{algorithmic}[1]                
\REQUIRE ~~\\                          
The resized image ${\mathcal I}$ with resolution $128\times128$
\ENSURE ~~\\                           
Saliency map ${\mathcal S}$ of ${\mathcal I}$ .
\vspace{0.15cm}
\STATE Convolve the input image with a Laplacian kernel, $ L$= [0 -1 0; -1 4 -1; 0 -1 0]. Obtain the gradient magnitude map $Gra$;
\STATE Convolve $Gra$ with a low-pass Gaussian filter kernel $g$, giving ${\mathcal S}=g{\star}|Gra|^2$;
\RETURN ${\mathcal S}$.                
\end{algorithmic}
\end{algorithm}

In the next section, we will discuss the question regarding whether the amplitude spectrum contains any useful information about saliency. We will illustrate  that the amplitude spectrum contains very important information and will develop a new framework for saliency detection in which we make full use of both the amplitude and phase.

\section{Convolution of the Amplitude Spectrum with a Low-Pass Gaussian Kernel Equals a Saliency Detector}
\label{SSS}

Many researchers have proposed models of saliency, which invariably then require the detection of salient {\it regions}. These regions are described as {\it distinctive }or {\it irregular} patterns, which possess a distinct feature distribution when compared with the rest of the image. In this paper, instead of searching for these irregular patterns, we model regular or  so-called common patterns that do not attract much attention by our visual system. We refer to these patterns as being {\it non-salient}.

\subsection{Suppressing Repeated Patterns for Saliency Pop-Out}

\begin{figure}[t]
\begin{center}

   \includegraphics[width=.7\linewidth]{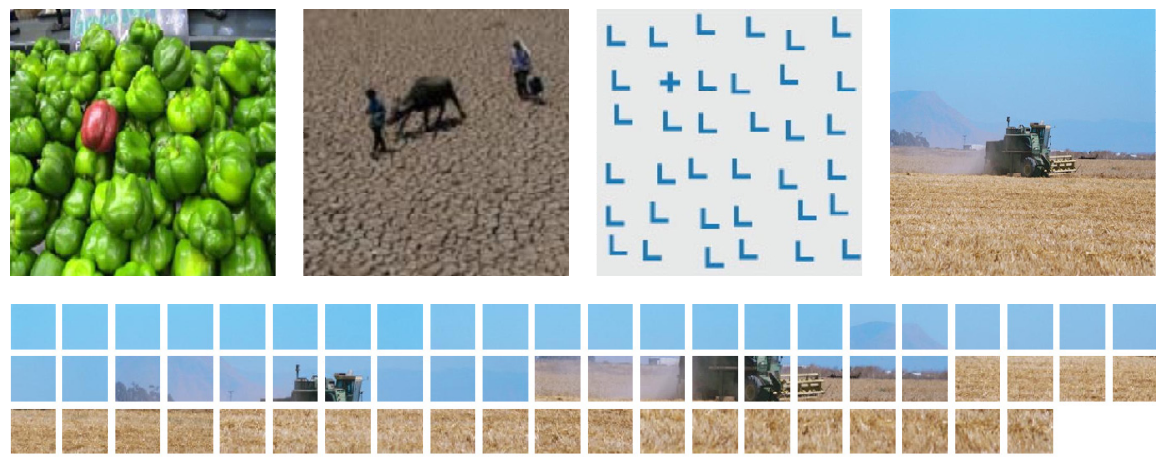}
\end{center}
   \caption{Regular(repeated) and anomalous patterns. Top: Four images; bottom: Collection of fragments from the last image above.}
\label{fig:repeatPatt}
\end{figure}

In the proposed model, we assume that a natural image consists of several salient and many so-called regular regions. All of these entities (whether distinct or not) may be considered as visual stimuli that compete for attention in the visual cortex. In this regard, it has been shown that nearby neurons  constituting receptive fields in the visual cortex mutually inhibit each other and interact competitively \cite{duncan1989visual}. As an example, in Fig. \ref{fig:repeatPatt}, if we divide the image into many patches (at a particular scale), we find that, some  are distinctive, while others are quite similar to each other. The bottom part of  Fig. \ref{fig:repeatPatt} shows the collection of patches from the last natural image above. We observe that several patterns appear many times (e.g., blue sky and grassy patches). We refer to these regular patches  as {\it repeated patterns}, which correspond to {\it non-saliency}.

Clearly, the primate visual system is more sensitive to distinctive rather than repeated patterns in an image.  Furthermore, the latter are very diverse. For example, consider the top row of Fig. \ref{fig:repeatPatt}. These exhibit several different examples of repeated patterns at different scales (including at the "scale" of 0 frequency for the uniform areas): grassy and sky patches (image 4), similar objects (image 1), road patches of the same color and texture (image 2), the 'L's (image 3), and so on. We model these repeated patterns and then suppress them, thereby producing the pop-out of the salient objects.

\subsection{Spikes in the Amplitude Spectrum Correspond to Repeated Patterns}

In this paper, we will illustrate that the amplitude spectrum contains important information corresponding to saliency and non-saliency. To be more precise, the spikes in the amplitude spectrum turn out to correspond to repeated patterns, which should be suppressed for saliency detection.

For convenience, we take a 1-D periodic signal $f(t)$ as an example. Suppose $f(t)$  can be represented by $f(t)=\sum_{n=-\infty}^{\infty} F(n)e^{jn\omega_1t}$,  where $F_n=\frac{1}{T}\int_{-T/2}^{T/2}f(t)e^{-jn\omega_1t}dt$. Then the Fourier transform is given by:
\begin{equation}
{\mathcal F}(w)=2\pi\sum_{n=-\infty}^{\infty} F(n)\delta(\omega-n\omega_1).  \label {equ13}
\end {equation}
From (\ref{equ13}), we can conclude that the spectrum of a periodic signal (repeated cycles) is a set of impulse functions (spikes). We note that this is based on the assumption that the signal is infinite. Therefore, given a more realistic finite length periodic signal, the shape of the spectrum will obviously be different but not degraded greatly.

Fig.\ref{fig:Spikes} provides an illustration of this point. Fig.\ref{fig:Spikes}(a) shows three signals with a different number of repeated patterns (cycles) while Fig.\ref{fig:Spikes}(b) shows their corresponding amplitude spectra. We observe that the larger the number of repeated cycles, the sharper the spikes in the spectrum. In order to quantify this notion, we define the {\it sharpness} of a spectrum $X$. We note that if we smooth the spikes by convolving the spectrum with a low-pass filter, the sharper the original spike, the more its peak height will be reduced. Therefore, the {\it sharpness} of $X$ can be defined as
${\mathcal \gamma}(X)= \parallel X-X{\star}h_{m}\parallel_\infty,$
where $h_{m}$ is a Gaussian kernel with fixed scale. The sharpness values of these three spectra in Fig. \ref{fig:Spikes} are 0.2320, 0.6091 and 1.3227 respectively.
Besides the sinusoid shown in the figure, other repeated signals also have this characteristic.

Next suppose there is one salient part that is embedded in a finite length periodic signal (row 1 of Fig. \ref{fig: ssf}). We will illustrate that this salient interval will not largely influence the spikes in the spectrum. That is to say, 1) The spikes will remain even though a salient part is embedded in the periodic signal; 2) The embedded salient part will not lead to very sharp spikes in the spectrum. The signal to be analyzed is defined as follows:
\begin{equation}
 f(t)=g(t)+g_\tau(t)+s(t), \label {equ15}
\end {equation}
where $g(t)$ is a periodic signal with finite length $L$, equaling $p(t)$ inside the interval $(0, L)$ and 0 elsewhere;
$g_\tau(t)=-p(t)\cdot {r}(t)$;
$s(t)=p_s(t)\cdot {r}(t)$,
$s(t)$ is the salient part of $f(t)$, which for convenience is also defined as a portion of yet another periodic function $p_s(t) $; $p(t)$ and $p_s(t) $ are periodic functions with frequencies $\nu$ and $\nu_s$, respectively; ${r}(t)$ is a rectangular window function that equals 1 inside the interval $(t_0,t_0+\tau)$ and 0 elsewhere; we also suppose that $(t_0,t_0+\tau)\in (0,L)$ and  $\tau \ll L$ (see row 1 of Fig. \ref{fig: ssf}). Thus the Fourier Transform of $f(t)$ can be represented as follows:
\begin{multline}
{\mathcal F}(f)(\omega)=\int_{-\infty}^{\infty}f(t)e^{-j\omega t}dt=
\int_{0}^{L}g(t)e^{-j\omega t}dt\\
+\int_{ t_0}^{ t_0+\tau}g_\tau(t)e^{-j\omega t}dt+\int_{ t_0}^{ t_0+\tau}s(t)e^{-j\omega t}dt\label{F(f)}.
\end{multline}
From (\ref{F(f)}), the spectrum of $f(t)$ consists of three terms. We assume that $\tau \ll L$. This implies that the first term has very sharp spikes in the amplitude spectrum as it contains many repeated patterns, while this is not true of the second and third terms. Consider $g_\tau(t)$ as an example. $g_\tau(t)$  is the point-wise product of signal $-p(t)$ and ${r}(t)$. According to the convolution theorem, ${\mathcal F}(g_\tau)(\omega)$ equals the convolution of $-{\mathcal F}(p)(\omega)$ with ${\mathcal F}(r)(\omega)$. Since ${\mathcal F}({r})(\omega)=\frac{2sin (\tau/2)}{\omega}e^{j\omega(t_0+\tau/2)}$ is a low-pass filter, the spikes in the amplitude spectrum of $-{\mathcal F}(p)(\omega)$ will be greatly suppressed. That is to say, there are no sharp spikes in the second term. This also occurs for the third term.
As discussed above, the sharpness of ${\mathcal F}(f)(\omega)$ is mainly determined by $g(t)$, while the latter two terms in (\ref{F(f)}) do not make a significant contribution to the spikes in the spectrum. In other words, since the first term corresponds to repeated patterns (non-salient) which lead to spikes, they can be suppressed by smoothing the spikes in the amplitude spectrum of ${\mathcal F}(f)(\omega)$.

\begin{figure}[t]
\begin{center}
   \includegraphics[width=.9\linewidth]{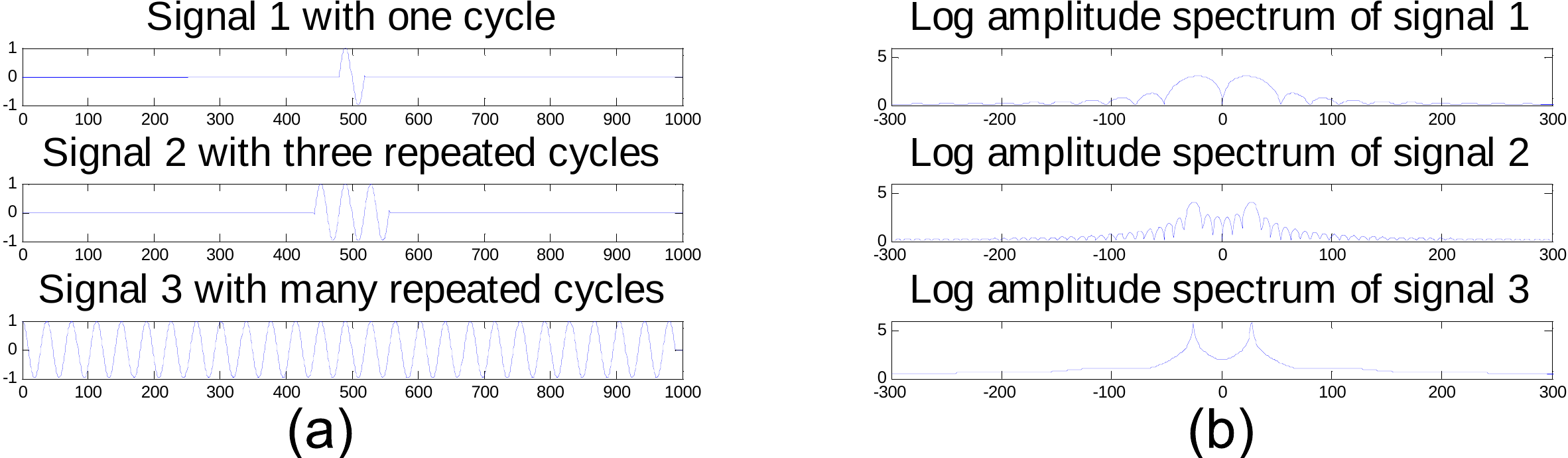}
\end{center}
   \caption{Repeated patterns  lead to sharp spikes: (a): Signals with different number of repeated cycles; (b): corresponding amplitude spectra.}
\label{fig:Spikes}
\end{figure}
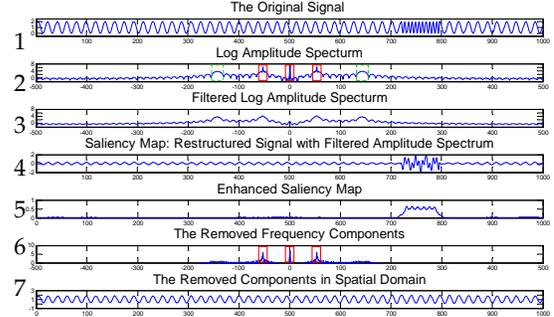
\begin{figure}[h]
\begin{center}
\begin{overpic}[width=7cm]{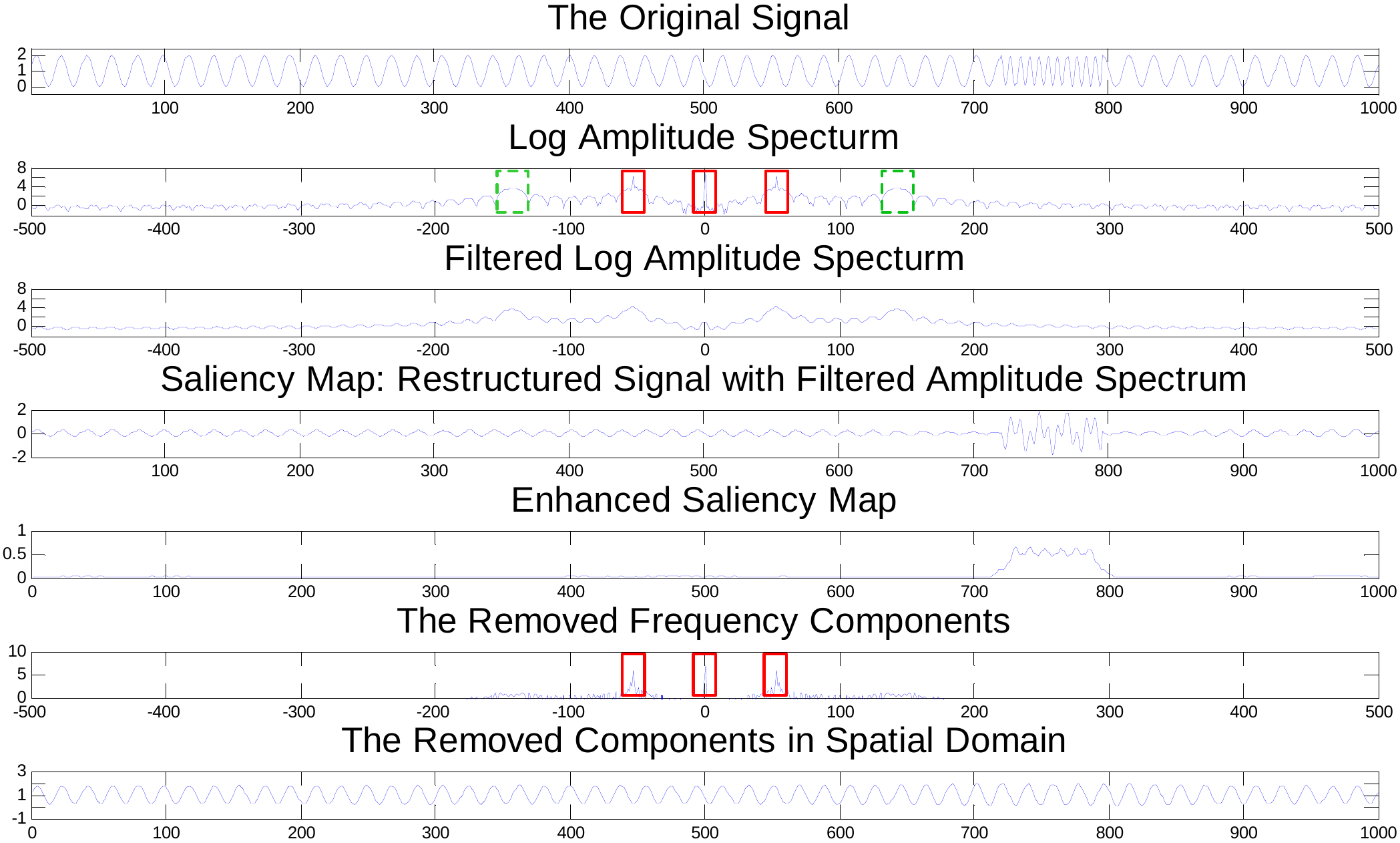}
\put(-2,51) {{\small 1}}
\put(-2,43) {{\small 2}}
\put(-2,35) {{\small 3}}
\put(-2,27) {{\small 4}}
\put(-2,19) {{\small 5}}
\put(-2,12) {{\small 6}}
\put(-2,4) {{\small  7}}
\end{overpic}
\caption{Suppression of repeated patterns by using spectrum filtering. It is clear that the larger the repeated background, the sharper the spikes, leading to the suppression of the amplitude spectrum via filtering.}
\label{fig: ssf}
\end{center}
\end{figure}

\subsection{Suppressing Repeated Patterns Using Spectral Filtering}

A Gaussian kernel $h$ can be employed to suppress spikes in the amplitude spectrum $|{\mathcal F}\{f\}|$ of an image as follows\footnote {In the implementation of this equation, we found that suppressing spikes in the log amplitude spectrum rather than the amplitude spectrum yielded better results.}:
\begin{equation}
\mathcal{A_{S}}(u,v)=|{\mathcal F}\{f(x,y)\}|\star h.
\label{Eq:AmpSmth1}
\end{equation}
The resulting smoothed amplitude spectrum ${\mathcal{A}_{S}}$ and the {\it original} phase spectrum are combined to compute the inverse transform, which in turn, yields the saliency map:
\begin{equation}
{\mathcal S}={\mathcal F^{-1}}\{\mathcal{A_{S}}{(u,v)}e^{i\cdot\mathcal{P}{(u,v)}}\}.
\label{Eq:AmpSmth2}
\end{equation}
In order to improve the visual display of saliency, we define it hereafter as:
\begin{equation}
{\mathcal S}=g\star|{\mathcal F^{-1}}\{\mathcal{A_{S}}{(u,v)}e^{i\cdot\mathcal{P}{(u,v)}}\}|^{2}.
\label{Eq:AmpSmth3}
\end{equation}

Consider the very simple example shown in Fig.\ref{fig: ssf}. The input signal (row 1) is periodic, but there is a short segment for which a different frequency signal is apparent. The short segment is quite distinct from the background for human vision, so a saliency detector should be able to highlight it. Row 2 shows the amplitude spectrum: there are three very sharp spikes (labeled by solid boxes), one of which corresponds to the constant background (uniform part) at zero frequency and the other two correspond to the periodic background. In addition, there are two rounded maxima (labeled by dashed boxes) corresponding to the salient parts. The amplitude spectrum is then smoothed by a Gaussian kernel (row 3), and the signal is reconstructed using the smoothed amplitude and original phase spectrum (row 4). It is clear that both the periodic and the uniform background are largely suppressed while the salient segment is well preserved. Row 5 shows the saliency map after enhancing the signal shown in row 4 using post-processing. We can further analyze this in the frequency domain, as shown in row 6, which illustrates the components actually removed by the previous operations. Here the eliminated frequency components are mainly the low frequencies near zero frequency, as well as the periodic background. Row 7 presents these removed components in the spatial domain. We find that non-salient parts (including uniform parts) are well suppressed using amplitude filtering. This process suggests that convolution in the frequency domain of the amplitude spectrum with a Gaussian kernel is equivalent to an image saliency detector\footnote{One might mistakenly be confused to think that this convolution in the frequency domain is equivalent to multiplication in the spatial domain as in the convolution theory. Yet, this is not the case as we convolve only the amplitude and do not change the phase.}.

\subsection{Spectrum Scale-Space Analysis}
\label{sec:SSSA}

\begin{figure}
\begin{center}
   \includegraphics[width=.9\linewidth]{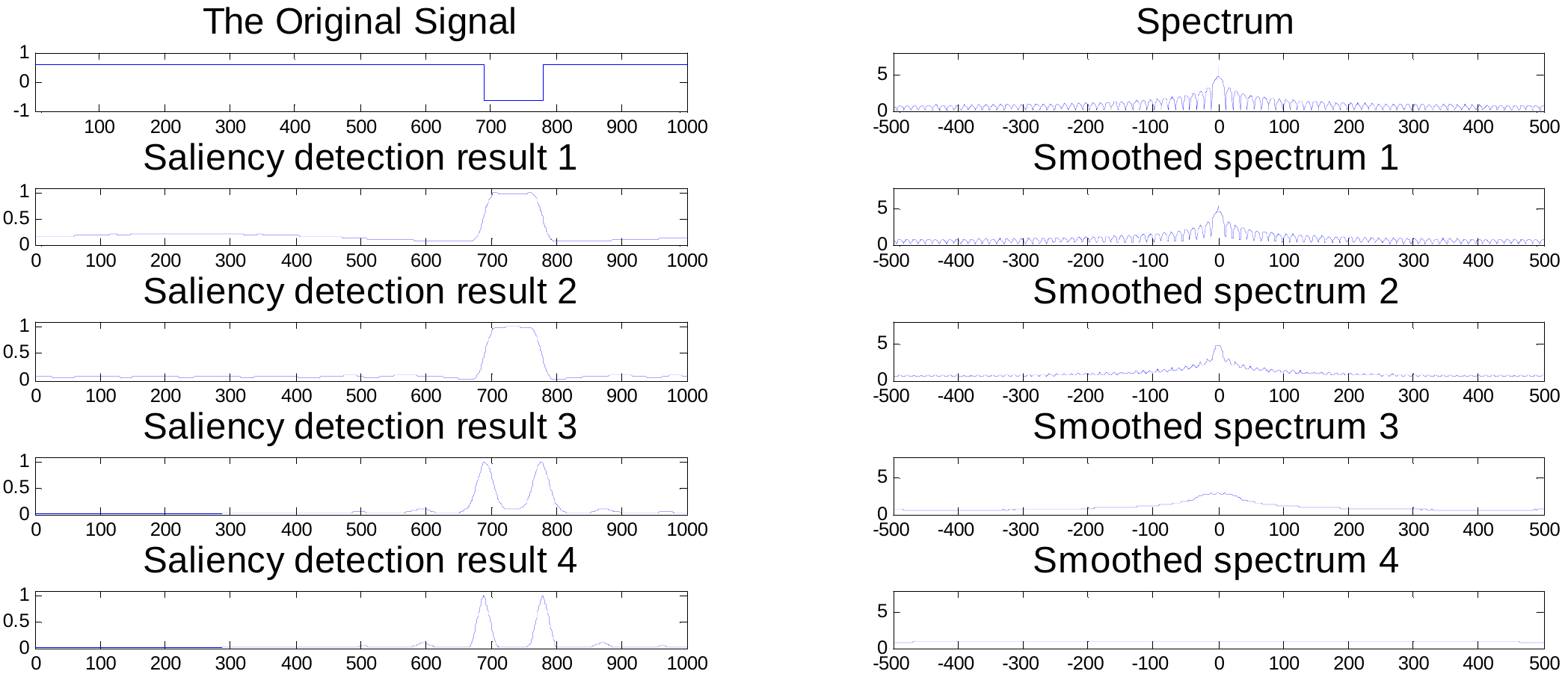}
\end{center}
   \caption{The original 1D signal is shown in the first row of col. 1 with the 1D saliency maps below. The spectrum of the original signal is shown in the first row of col. 2, followed by the smoothed spectra associated with the paired saliency map in col. 1.}
\label{fig:Scale}
\end{figure}

Repeated patterns (including uniform patterns) can be suppressed by smoothing the amplitude spectrum at an appropriate scale. However, which scale is the best in (\ref{Eq:AmpSmth1})? As shown in Fig. \ref{fig:Scale}, if the filter scale is too small, the repeated patterns cannot be suppressed sufficiently (row 2), while if the filter scale is too large, only the boundaries of the salient region are highlighted (row 4 and 5). Therefore it is important to select a proper scale for the Gaussian kernel. In fact, we will illustrate that different filter scales are required for different types of saliency. For example, a small-scale kernel is needed to detect large salient regions, while a large-scale kernel could be used to detect texture-rich or small salient regions (e.g. distant objects in the scene).

In this paper, we propose a Spectrum Scale-Space (SSS) for handling amplitude spectra at different scales, yielding a one-parameter family of smoothed spectra which is parameterized by the scale of the Gaussian kernel. Given an amplitude spectrum, $\mathcal{A}(u,v)$, of an image, the SSS is a family of derived signals $\Lambda(u,v;k)$ defined by the convolution of $\mathcal{A}$ with the series of Gaussian kernels:
\begin{equation}
g(u,v;k)=\frac{1}{\sqrt{2\pi}2^{k-1}t_0}e^{-(u^2+v^2)/(2^{2k-1}t_0^2)},
\label{Eq:GassianKernel}
\end{equation}
where $k$ is the scale parameter, $k=1,...,K$. $K$ is determined by the image size: $K=\lceil log_2{min\{H,W\}}\rceil+1$, where $H$, $W$ indicate the height and width of the image; $t_0=0.5$. Thus the scale space is defined as follows:
\begin{equation}
\Lambda(u,v;k)=(g(.,.;k)\star\mathcal{A})(u,v).
\label{Eq:SSS}
\end{equation}

\begin{figure}[h]
\begin{center}
\begin{overpic}[width=7.5cm]{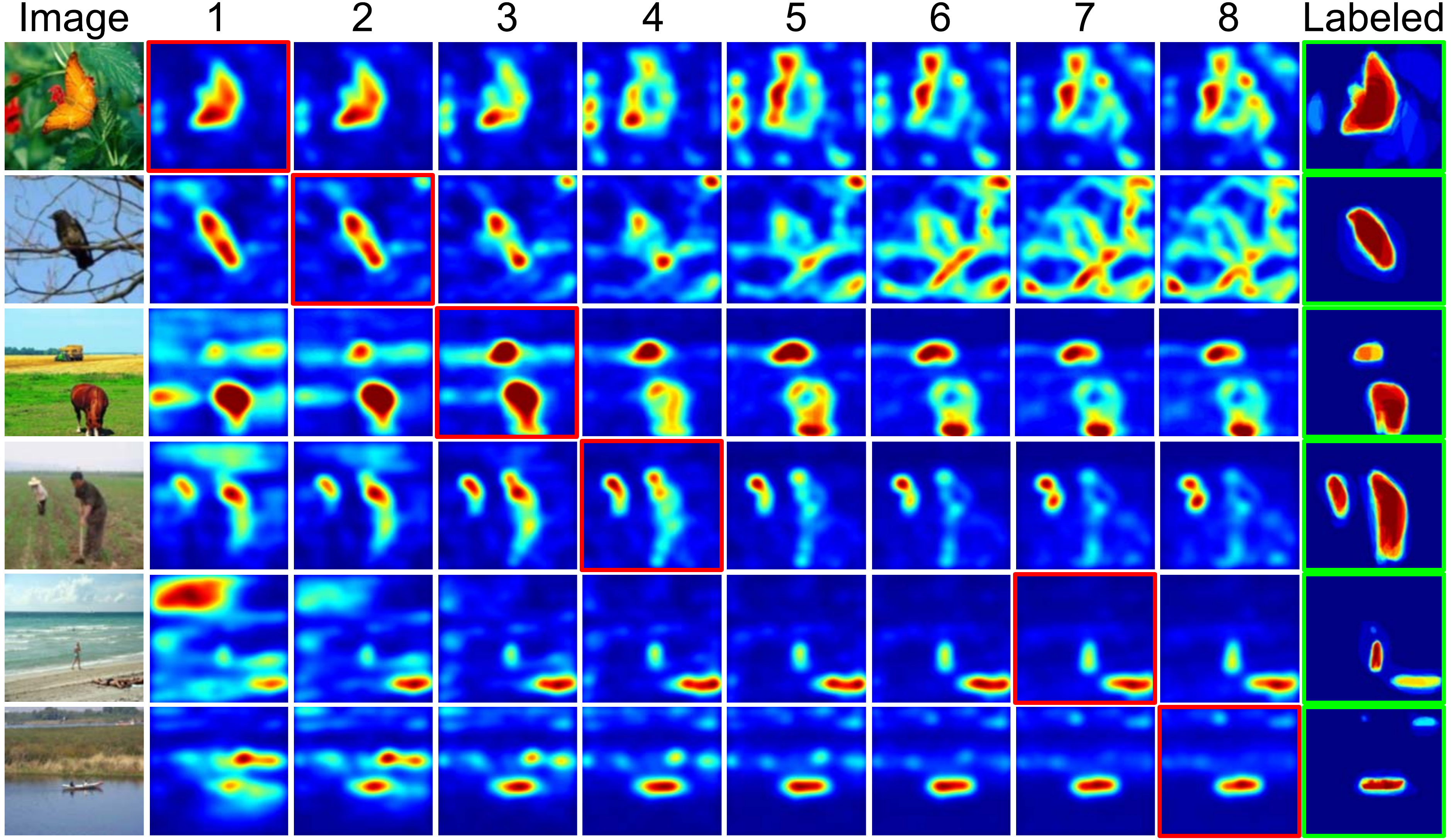}

\end{overpic}
\caption{Five 2-D examples are shown. The first column shows the original 2-D signals (images). The remaining images in each row present the set of saliency maps computed by smoothing the original image amplitude spectrum using different scales for the Gaussian kernels.}
\label{fig:Scale2}
\end{center}
\end{figure}
As an example, assume a 1-D signal. We first compute a series of filtered spectra according to the SSS model; then the saliency map is computed for each scale, as shown in Fig. \ref{fig:Scale}. The significance of the scale for saliency detection can easily be observed. In this example, smoothed spectrum 2 gives the best result. As the kernel scale goes to infinity, the spectrum tends to be a constant (horizontal plane in 2D), as shown in the last row of Fig. \ref{fig:Scale}. This is exactly the case proposed in \cite{hou2007saliency, guo2008spatio, guo2010multiresolution}.

Fig. \ref{fig:Scale2} shows 2D results obtained using  different kernel scales, increasing from left to right. The best saliency map is labeled by a red square. We observe that broad regions pop out when smaller scale kernels are used, while distant objects or those with rich texture pop out when larger scale kernels are used. Thus given a natural image, a set of saliency maps is obtained from which one must be selected as the final saliency map. The criterion for achieving this will be discussed in section \ref{sM:HFT}.

Here, we suggest that the frequency-tuned model \cite{LCAV-CONF-2009-012} is, to some extent, a special case of the proposed model. In \cite{LCAV-CONF-2009-012}, the saliency map is defined as:
${\mathcal{S}}(x,y)=\Vert I_{\mu}-I_{\omega hc}(x,y)\Vert$,
where $I_{\mu}$ is the average $Lab$ vector of the entire image and $I_{\omega hc}(x,y)$ is a specific $Lab$ pixel vector from the Gaussian-filtered version of the original image. Authors compute a saliency map by removing the frequencies around the DC frequency (the {"}mean{"}). Previously, we have illustrated that there is always a very sharp spike around zero frequency, which corresponds to this {"}mean{"}. Hence, if we use a very small scale Gaussian kernel to smooth the spectrum, those components corresponding to the {"}mean{"} will be suppressed significantly.

\section{Saliency Using the  Hypercomplex Fourier Transform (HFT)}
\label{sM:HFT}

In section \ref{SSS}, we discussed the saliency computation using only one feature map (that of intensity). However, in order to obtain better performance, more features are required, for example, color and motion information. Inspired by \cite{guo2008spatio, ell2002quaternion, ell2006hypercomplex}, we use the so-called hypercomplex matrix to combine multiple feature maps. Consequently, the {\it Hypercomplex Fourier Transform} (HFT) is employed to replace the {\it Fourier Transform} used in section \ref{SSS} for saliency computing.

\subsection{Hypercomplex Fourier Transform}

The input to the traditional Discrete Fourier Transform is a real matrix. Each image pixel is an element of the input matrix and is a real number. However, if we combine more than one feature into a hypercomplex matrix, each element is a vector and this hypercomplex matrix is a {\it vector field}. Thus, the traditional Fourier Transform becomes unsuitable for computational purposes.

The Hypercomplex Fourier Transform was proposed in \cite{ell2002quaternion}, in which the hypercomplex input was specified to be a quaternion\footnote{The quaternion is represented as $q=a+bi+cj+dk$, where a,b,c and d are real numbers and i,j,k satisfy $ i^{2}=j^{2}=k^{2}=ijk=-1$. A quaternion can also be represented as $q=S(q)+V(q)$, where $S(q)=a$ is the scalar and $V(q)=bi+cj+dk$ is the vector part. q is called a pure quaternion if $S(q)=0$.}.  Given a hypercomplex matrix:
\begin{equation}
f(n,m)=a+bi+cj+dk,
\label{Eq:qft1}
\end{equation}
the discrete version of the HFT of (\ref{Eq:qft1}) is given by:
\begin{equation}
\mathcal {F_{H}}[u,v]=\frac{1}{\sqrt{MN}}\sum^{M-1}_{m=0}{\sum^{N-1}_{n=0}e^{-\mu 2\pi((\frac{mv}{M})+(\frac{nu}{N}))}}f(n,m),
\label{Eq:qft2}
\end{equation}
where $\mu$  is a unit pure quaternion and $\mu^{2}$ =-1. Note that $\mathcal {F_{H}}[u,v]$ is also a hypercomplex matrix. The inverse Hypercomplex Fourier Transform is given as:
\begin{equation}
f(n,m)=\frac{1}{\sqrt{MN}}\sum^{M-1}_{v=0}{\sum^{N-1}_{u=0}e^{\mu 2\pi((\frac{mv}{M})+(\frac{nu}{N}))}}\mathcal {F_{H}}[u,v].
\label{Eq:qft3}
\end{equation}

\subsection{Hypercomplex Representation of Multiple Feature Maps}
The Hypercomplex representation can be employed to combine multiple features (e.g., in \cite{guo2010multiresolution} the authors combine color, intensity and motion as the features). We define the input hypercomplex matrix as follows:
\begin{equation}
f(n,m)=w_{1}f_{1}+w_{2}f_{2}i+w_{3}f_{3}j+w_{4}f_{4}k,
\label{Eq:qft5}
\end{equation}
where ${w_{1}\text{-}w_{4}}$  are weights and ${f_{1}\text{-}f_{4}}$  are the feature maps (matrices). Similar to \cite{Itti_etal98pami}, we use three features to compute the saliency for the static input case:
\begin{equation}
f_{2}=\mathcal {I}_{s}=(r+g+b)/3,
\label{Eq:fm1}
\end{equation}
\begin{equation}
\label{Eq:fm2}
f_{3}=\mathcal {RG}=R-G,
\end{equation}
\begin{equation}
\label{Eq:fm3}
f_{4}=\mathcal {BY}=B-Y,
\end{equation}
where $r$, $g$, $b$ are the red, green and blue channels of an input color image and $R=r-(g+b)/2$, $G=g-(r+b)/2$, $B=b-(r+g)/2$, $Y=(r+g)/2-|r-g|/2-b$.
These three feature maps comprise the opponent color space representation of the input image (see part 1 of Fig. \ref{fig:pHFT}). Based on the work in \cite{guo2010multiresolution}, our approach has also been experimentally confirmed using videos by defining a motion feature $\mathcal M$ and setting $f_{1}=\mathcal M$  in (\ref{Eq:qft5}). In this paper, we consider only the static image case by employing just intensity and color information. We select the weights so that $w_{1}=0$,  $w_{2}=0.5$,  $w_{3}=w_{4}=0.25$.

\subsection{Computing the Saliency Map}

\begin{figure}
\begin{center}
   \includegraphics[width=1\linewidth]{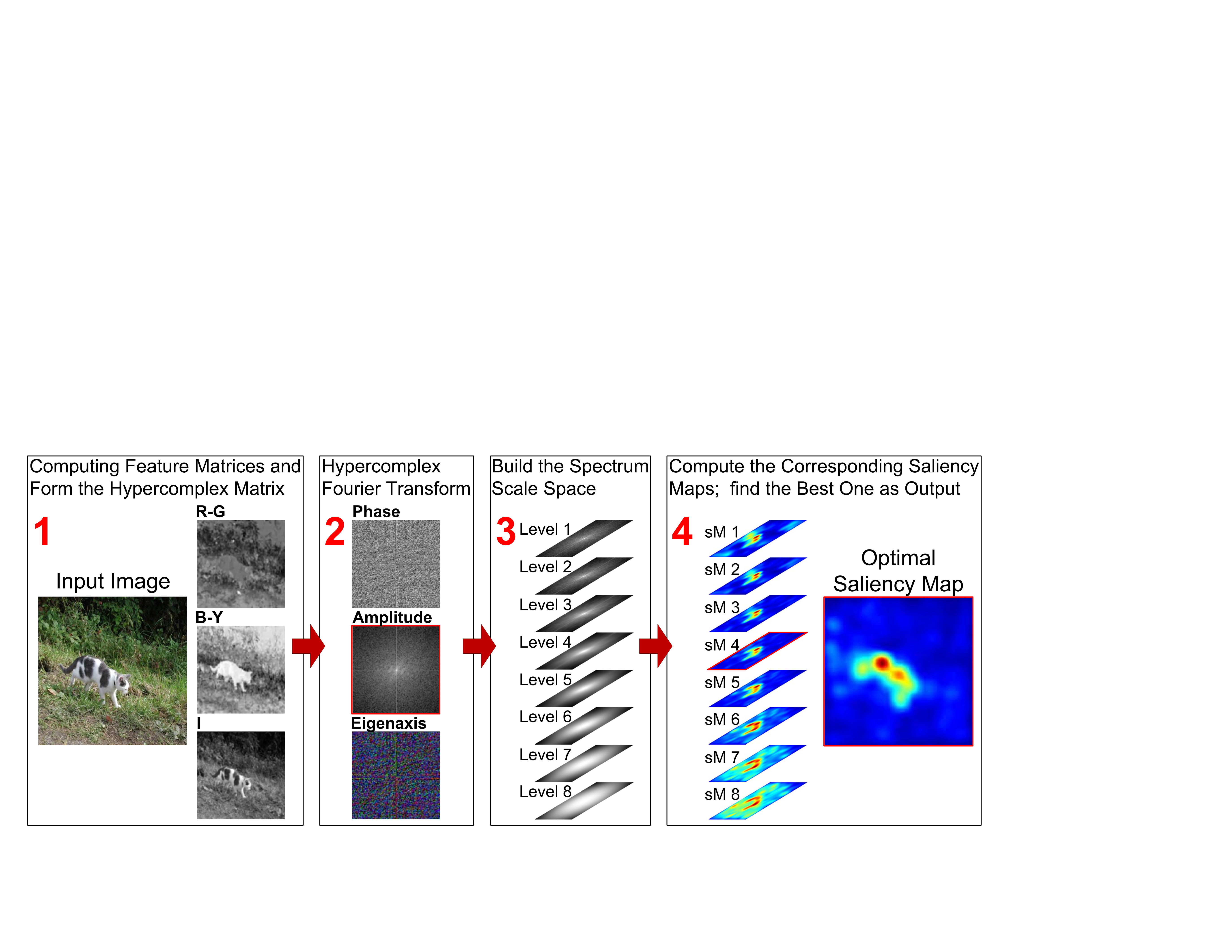}
\end{center}
   \caption{Procedure for computing  saliency  using the Hypercomplex Fourier Transform (HFT)}
\label{fig:pHFT}
\end{figure}

Given an image, the input is defined according to section 4.2. The Hypercomplex Fourier Transform, $\mathcal {F_{H}}[u,v]$, can be rewritten in polar form as follows:
\begin{equation}
\mathcal {F_{H}}[u,v]=\Vert\mathcal {F_{H}}[u,v]\Vert e^{\mu \Phi (u,v)},
\label{Eq:qft6}
\end{equation}
where $\Vert\cdot \Vert$ indicates the modulus for each element of a hypercomplex matrix;
$\mathcal {F_{H}}[u,v]$ can be considered as the frequency domain representation of $f(m,n)$.  Its amplitude spectrum $\mathcal{A}(u,v)$, phase spectrum $\mathcal{P}(u,v)$ and the so-called eigenaxis spectrum $\mathcal{X}(u,v)$ are defined as:
$$\mathcal{A}(u,v)=\Vert\mathcal{F_{H}}(u,v)\Vert,$$
$$\mathcal{P}(u,v)={\Phi}{(u,v)}=tan^{-1}\frac{\Vert\mathcal{V}{(\mathcal{F}(u,v))}\Vert}{\mathcal{S}{(\mathcal{F}(u,v))}},$$
$$\mathcal{X}(u,v)=\mu (u,v)=\frac{\mathcal{V}({\mathcal{F}(u,v)})}{\Vert \mathcal{V}({\mathcal{F}(u,v)})\Vert},$$
where, $\mathcal{X}(u,v)$ is a pure quaternion matrix. These three spectra are shown in part 2 of Fig. \ref{fig:pHFT} \footnote{Here we use a monochrome image to represent the phase spectrum $\mathcal{P}(u,v)$ as it is a real matrix. This is different from  \cite{ell2006hypercomplex}.}.

As discussed in Sec. \ref{SSS}, the amplitude spectrum contains important information about the scene. Similar to the discussion in sec. \ref{sec:SSSA}, we create the {\it Spectrum Scale Space} $\Lambda=\{\Lambda_{k}\}$ by smoothing $\mathcal{A}(u,v)$ with a series of Gaussian kernels according to (\ref{Eq:SSS}) (see Fig.\ref{fig:pHFT} (3)) while retaining unchanged the phase spectrum $\mathcal{P}(u,v)$ and eigenaxis spectrum $\mathcal{X}(u,v)$.

Observing the images in part 3 of Fig. \ref{fig:pHFT} reveals that when the scale $k$ is very small, the information contained in the amplitude plots is retained quite well, while when it becomes very large, the pertinent information basically is lost. Actually, the PQFT model is a special case of the proposed framework when the scale goes to infinity.

Thus, given a single (smoothed) amplitude spectrum $\Lambda_{k}$ (one layer in $\Lambda$) and the original phase and eigenaxis spectra, we can perform the inverse transform (\ref{Eq:qft3}) to give the saliency map at each scale:
\begin{equation}
{\mathcal{S}_{k}}=g\star\Vert\mathcal{F}_{\mathcal{H}}^{-1}\{\Lambda_{k}(u,v)e^{\mathcal{X} \mathcal{P} (u,v)}\}\Vert^{2},
\label{Eq:qft7}
\end{equation}
where $g$ is a Gaussian kernel at a fixed scale\footnote{For convenience, the scale parameter has been set to $0.05\cdot W$. Though this has been done to improve the visual display, it will nevertheless influence the ROC score when predicting human fixation \cite{ImageSignature2012}. We discuss this issue in detail in section \ref{Sec:Experiment}.}. Thus, we again obtain a series of saliency maps $\{\mathcal{S}_{k}\}$, shown in part 4 of Fig.\ref{fig:pHFT}. In the  approach proposed in this paper, the final saliency map $\mathcal{S}$ is chosen from $\{\mathcal{S}_{k}\}$ by selecting the best scale $k_{p}$ according to criteria discussed in section \ref{findingscale}. The saliency model based on the  Hypercomplex Fourier Transform is referred to as HFT in this paper.

The HFT model is  summarized in Algorithm \ref{alg:HFT} \footnote{The input image is resized to $128 \times 128$ in the experiments.}.
\begin{algorithm}[htb]         
\caption{{\bf HFT }saliency model}             
\label{alg:HFT}                  
\begin{algorithmic}[1]                
\REQUIRE ~~\\                          
The resized color image ${\mathcal{C}}$ with resolution $m\times n$
\ENSURE ~~\\                           
Saliency map ${\mathcal{S}}$ of ${\mathcal{C}}$
\vspace{0.15cm}
\STATE Compute the feature maps $\{\mathcal{I}$, $\mathcal{RG}$, $\mathcal{BY}\}$ of ${\mathcal{C}}$ according to (\ref{Eq:fm1})-(\ref{Eq:fm3});
\STATE Form the hypercomplex matrix $f(n,m)$ by combining these feature maps according to (\ref{Eq:qft5});
\STATE Perform the Hypercomplex Fourier Transform on $f(n,m)$ and compute the amplitude spectrum $\mathcal{A}$, phase spectrum, $\mathcal{P}$ and eigenaxis spectrum $\mathcal{X}$;
\STATE Smooth the amplitude spectrum with Gaussian kernels according to (\ref{Eq:GassianKernel}), thereby obtaining a spectrum scale space  $\{\Lambda_{k}\}$;
\STATE Obtain a saliency map ${\mathcal{S}}_{k}$ according to (\ref{Eq:qft7}) for each $\Lambda_{k}$, thereby producing a sequence of saliency maps $\{{\mathcal{S}}_{k}\}$;
\STATE Find the best saliency map ${\mathcal{S}}$ from the set $\{{\mathcal{S}}_{k}\}$ and use it as the final saliency map  according to the criterion introduced in (\ref{Eq:pscal5}) in section \ref{findingscale};
\RETURN ${\mathcal{S}}$.                
\end{algorithmic}
\end{algorithm}

\subsection{Finding the Proper Scale}
\label{findingscale}

In section \ref{SSS}, we assumed that the best saliency map would appear at a specific scale in the sequence $\{{\mathcal{S}}_{k}\}$. Unlike the use of {\it entropy} in \cite{kadir2001saliency}, we employ it as the criterion for determining the optimal scale:
\begin{equation}
k_{p}=\underset{k}{\arg\min}\{\mathcal{H}\left(\mathcal{S}_k\right)\},
\label{Eq:pscal0}
\end{equation}
where $\mathcal{H}(x)=-\sum_{i=1}^{n}p_{i}\mathrm{log}p_{i}$ is the definition of entropy of $x$. The reason for using entropy is as follows. The saliency map can be considered as a probability map. In a desirable saliency map, the regions of interest would be assigned higher values and the rest of the map would be largely suppressed. Thus it is expected that the values in the saliency map histogram  would cluster around certain values. The entropy of the saliency map would then be very small according to the definition of entropy.

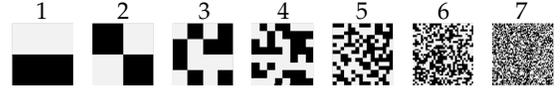
\begin{figure}[h]
\begin{center}
\begin{overpic}[width=7.3cm]{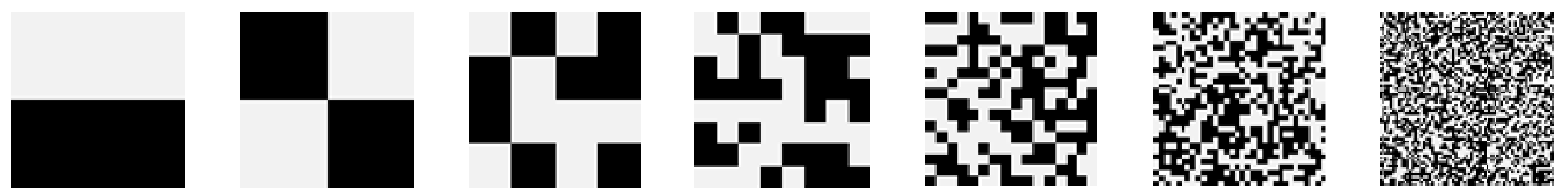}
\put(5,12) {{\small 1}}
\put(19.8,12) {{\small 2}}
\put(34.6,12) {{\small 3}}
\put(49,12) {{\small 4}}
\put(63.3,12) {{\small 5}}
\put(78.2,12) {{\small 6}}
\put(92.4,12) {{\small  7}}
\end{overpic}
\caption{Binary images with the same histogram, but with different spatial structures}
\label{fig:2dentropy}
\end{center}
\end{figure}

Conventional entropy is based on the distribution of a variable $x$; if the histogram is given, the entropy of $x$ is determined. However, the spatial geometric information is ignored. As shown in Fig. \ref{fig:2dentropy}, images may possess the same histograms and therefore have the same entropy values, even though the spatial structure becomes more and more chaotic. Obviously, in saliency detection, we wish to avoid selecting a map with a high level of chaos.
Spatial geometric information needs to be considered in 2D signal analysis, and work related to this issue has been reported, such as the so-called 2D Entropy \cite{abutaleb1989automatic, chen1994fast}. Here we present a simple improved  definition of entropy in order to make use of the spatial geometric information. We consider each pixel individually and require it to also  depend on the values of its neighbors.  We achieve this objective by employing a Gaussian kernel to filter the 2D signal, and then compute the conventional entropy on the smoothed 2D signal. Consequently, the new entropy is defined as:
$\mathcal{H}_{2D}(x)=\mathcal{H}\{g_{n}\star x\}$
where $g_{n}$ is a low-pass Gaussian kernel with a scale of $\varsigma$.

\begin{figure}[h]
\begin{center}
\begin{overpic}[width=7cm]{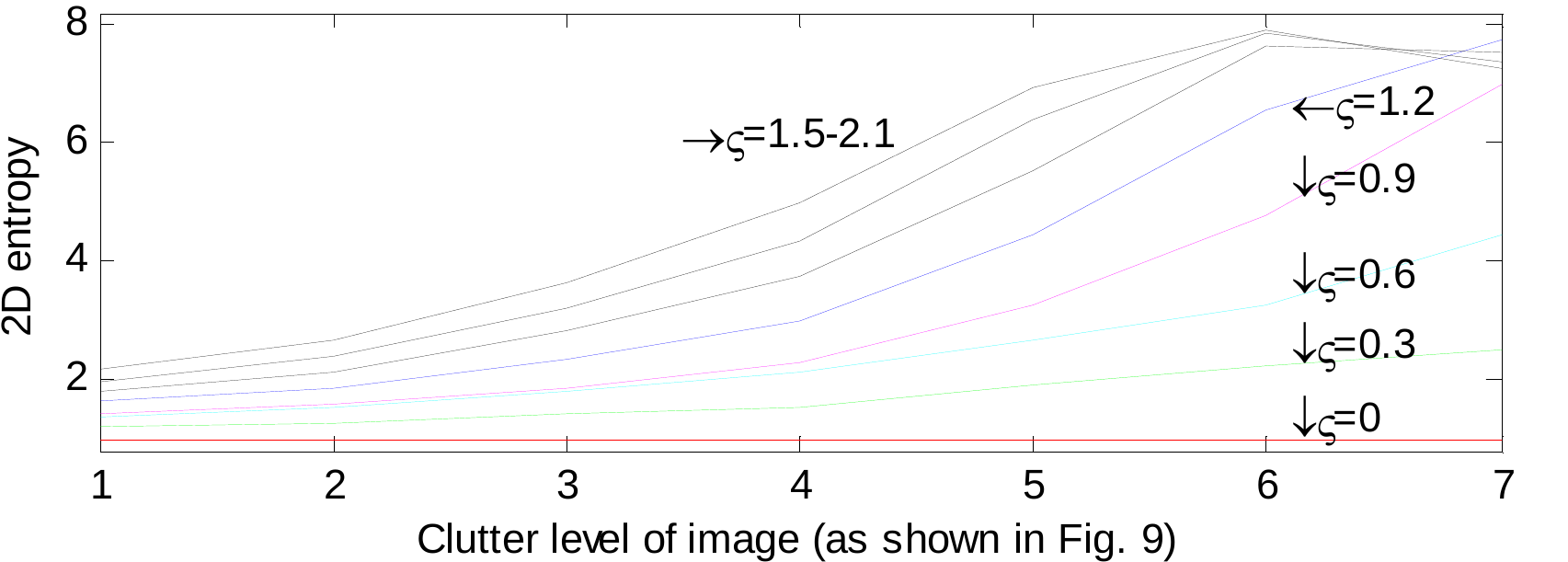}

\end{overpic}
\caption{Computing 2D entropy of the 2D images in Fig. \ref{fig:2dentropy} using Gaussian kernels of different size.}
\label{fig:2dentropy3}
\end{center}
\end{figure}

As shown in Fig. \ref{fig:2dentropy3}, if  $\varsigma$ were too small, especially when $\varsigma$=0, Gaussian filtering would have a minor effect. If $\varsigma$ =1.2, the entropy value would increase as the image became more and more chaotic in the spatial domain. This is quite reasonable. However, if $\varsigma$ were too large, the entropy value would decrease. This is because the small structures in the 2D signal would be heavily destroyed by the Gaussian filter. Thus, on the one hand, we desire that $\varsigma$ should be as large as possible, because with larger $\varsigma$ the influence of a pixel could spread farther. On the other hand, we do not wish to destroy the small spatial structures. Therefore, $\varsigma$ should be related to the size of smallest region we expect to detect. Experiments indicate that $\varsigma=0.01\thicksim0.03\cdot{W}$ yields acceptable results.

Besides entropy, there is another issue to consider when choosing the proper scale $k$. In HFT, given $\{{\mathcal{S}}_{k}\}$, we avoid choosing saliency maps with a strong response at the border region by using a border avoidance strategy. Thus, a parameter $\lambda$ is defined for each candidate saliency map:
$\lambda_{k}=\sum\sum{\mathcal{K}{}(n,m)\cdot\mathcal{N}(\mathcal{S}_{k}(n,m))},$
where $\mathcal{K}{}$ is a 2D centered Gaussian mask of the same size as ${\mathcal{S}}$, $\sigma_{w}=W/4$, $\sigma_{h}=H/4$, and $\sum\sum\mathcal{K}{}(n,m)=1$. $\mathcal{N}(\cdot)$ is used to normalize ${\mathcal{S}}$, so that the summation of all the pixel values is 1. Note that $\lambda$ is not the same as the {\it center-bias} or {\it border cut} described in \cite{zhang2008sun}, since it is used only to {\it choose} a proper scale, but not to {\it modify} the saliency map as done in \cite{NIPS2006_897,judd2012pami}. {\it Center-bias} and {\it border cut} will be discussed in section \ref{Sec:Experiment}.
Thus, with this definition of 2D entropy and $\lambda$, $k_{p}$ in (\ref{Eq:pscal0}) is revised as follows:
\begin{equation}
k_{p}=\underset{k}{\arg\min}\{\lambda_{k}^{-1}\mathcal{H}_{2D}\left(\mathcal{S}_k\right)\}.
\label{Eq:pscal5}
\end{equation}

The model that uses the criterion in the above equation is HFT. In addition, we use entropy without  the border-avoidance strategy as the criterion, and the results are labeled as HFT(e). Of course, it is possible that the performance of the proposed model might be improved with an even better criterion. For example, as shown in the row 2 of Fig.\ref{fig:Exp71}(F), HFT failed to highlight the two salient objects uniformly. However, we note that this was caused by the selection of an improper scale, notwithstanding the fact that the optimal scale did was present in the existing set, as shown in the third row of Fig.\ref{fig:Scale2}. In order to illustrate the potential power of the proposed model, we have also determined the optimal scale for each image visually by examining the ROC scores of the saliency maps. The results are reported in this paper and are labeled as HFT*.
\section{Experimental Results}
\label{Sec:Experiment}

Three experiments to evaluate the performance of the proposed HFT are discussed in this section: 1) Response to psychological patterns, 2) Predicting human fixations and 3) Predicting the object regions to which humans pay attention. Eight state-of-the-art methods were employed to perform the comparisons: Itti's model\cite{Itti_etal98pami}\footnote{This implementation comes in the GBVS package, see \url{http://www.klab.caltech.edu/~harel/share/gbvs.php}. The saliency toolbox STB (based on Itti's model) was used in the first experiment.}, DVA\cite{NIPS2008_0142}\footnote{The code is available at \url{http://www.its.caltech.edu/~xhou/}.}, GBVS\cite{NIPS2006_897}\footnote{ See \url{http://www.klab.caltech.edu/~harel/share/gbvs.php}.}, SR\cite{hou2007saliency}\footnote{The code is available at \url{http://www.its.caltech.edu/~xhou/}.}, PFT\cite{guo2008spatio}\footnote{PFT (Phase Fourier Transform) is an improved version of SR. Our implementation was done according to \cite{guo2008spatio}.}, PQFT\cite{guo2008spatio,guo2010multiresolution}\footnote{The code was provided by the first author in \cite{guo2010multiresolution}.}, AIM\cite{NIPS2005_81}\footnote{
See \url{http://www-sop.inria.fr/members/Neil.Bruce/AIM.zip}.} and SUN\cite{zhang2008sun}\footnote{The code is available at \url{http://cseweb.ucsd.edu/~l6zhang/}.}.

We evaluate the performance of saliency detection algorithms both qualitatively and by comparison to human observers. For the former, we essentially compare the saliency map to the original image by using a simple algorithm to determine an object map based on the saliency map. For the latter we require ground truth data. We use two kinds of ground truth in this paper, fixation data and salient regions labeled by human observers.
In section \ref{exp_fixation}, we have used freely available human fixation data \cite{NIPS2005_81} as ground truth to evaluate the algorithms listed earlier. ROC score (area under the ROC curve, AUC) is adopted to measure their performance.
In section \ref{exp_region}, we have also evaluated the algorithms using object regions labeled by humans (some examples of "labeled" results are shown in each second column of Fig.\ref{fig:Exp71}) as ground truth. In fact, the {\it available} eye tracking data only contain {\it positional} information \cite{hou2007saliency}. However, saliency detection algorithms in computer vision are assumed and expected to have the ability to detect salient object {\it regions} in a scene \cite{elazary2008salient}. For example, given a region such as a flower (see row 4 of Fig.\ref{fig:Exp71}(A) as an example), an algorithm should respond more or less uniformly within the whole region and not just along the boundary of the flower or at several points on the flower. Therefore, we use {\it salient region maps} labeled by humans as ground truth.
In this experiment, besides ROC, we also use the DSC (Dice Similarity Coefficient) as a measure to evaluate the overlap between the thresholded saliency map and the ground truth. The peak value of the DSC curve (PoDSC) is an important index of performance, as it corresponds to the optimal threshold and the best possible algorithm performance\cite{thomasevaluation2008}.

\subsection{How to Make Fair Quantitative Comparisons?}

There are two aspects which should be considered when making quantitative comparisons between two saliency models: scale and post-processing.

Certain models permit the usage of different image scales (input image size). Therefore, in these cases,  we find the optimal scale by maximizing their performance, but for the other models it is necessary to use the default settings, as shown in Table \ref{alg_list}.

With regard to post-processing, most previous work has used the ROC directly without investigating any of the post-processing factors affecting the fairness of this approach. However, it is important to note that three factors dramatically influence the ROC score and PoDSC: {\bf 1)} {\bf Border cut (BC)} \cite{zhang2008sun}, {\bf 2)} {\bf Centre-bias setting (CB)} \cite{NIPS2006_897} and {\bf 3)} {\bf smoothing (SM)} \cite{ImageSignature2012,NIPS2006_897}.
In this paper, in order to make a fair comparison, the post-processing is calibrated. We first consider BC and CB by dividing the saliency models into three {\it subsets}: 1) models without any BC and CB; 2) models with BC and 3) models with CB, as shown, in Table \ref{alg_list}. In addition, the optimal smoothing parameter for each model is learnt in order to eliminate the influence of SM. We compare HFT class models (HFT, HFT(e) and HFT*) with each of the the three subsets.

1) When comparing HFT class models with models in subset 1, we compute the ROC and/or PoDSC directly;

2) When comparing HFT class models with subset 2, we set the border cuts for all of these models to be of equal size\footnote{In our experiments, we  considered only the interior of the frame and the corresponding region in the ground truth when computing the ROC curve.}.

3) When comparing HFT class models with models in subset 3, we apply an {\it optimal} center-bias for each model individually, thereby ensuring that the ROC score for each model is maximized.

\begin{table}[htbp] 	
\caption{Three subsets of algorithms employed for comparison}
\label{alg_list}	
\centering 	
\begin{tabular}{@{}clcccc@{}}
\toprule
&\multicolumn{1}{c}{}
&\multicolumn{1}{c}{}
&\multicolumn{3}{c}{post-processing effects}\\
 \cmidrule(lr){4-6}
\scriptsize{}  {Subset} & {Model} &  {Image size}
& {SM} & {BC} & {CB}\\
\midrule
1 &HFT  &$128\times128^{\ddag}$ & explicit & no & no\\
1 &SR/PFT  &$64\times64^{\dag}$ & explicit & no & no\\
1 &SUN   &$\frac{1}{8}$full size$^{\dag}$& implicit &  no & no\\
\midrule
2 &AIM  &  $\frac{1}{2}$full size$^{\dag}$& implicit & yes & no\\
2 &DVA  & $80\times120^{\ddag}$ & explicit & yes & no\\
2 &PQFT  &$64\times64^{\dag}$& explicit & yes & no\\
2 &Itti  & full size$^{\ddag}$ & explicit  & yes & no\\
\midrule
3 &GBVS  & full size$^{\ddag}$ & explicit  & no & yes\\
\bottomrule 	
\end{tabular}\\[2pt]
    \footnotesize
    $^{\dag}$The optimal image size for this model. $^{\ddag}$The default image size.
\end{table}

\subsection{Response to Psychological Patterns}
\label{EXP:Psy}
Psychological patterns are employed to evaluate three aspects of the algorithms: 1) first we use them to evaluate basic detection ability; 2) then we evaluate their ability to detect salient regions of different sizes, and 3) we evaluate their tolerance to random noise.

\begin{figure}[h]
\begin{center}
\begin{overpic}[width=7.2cm]{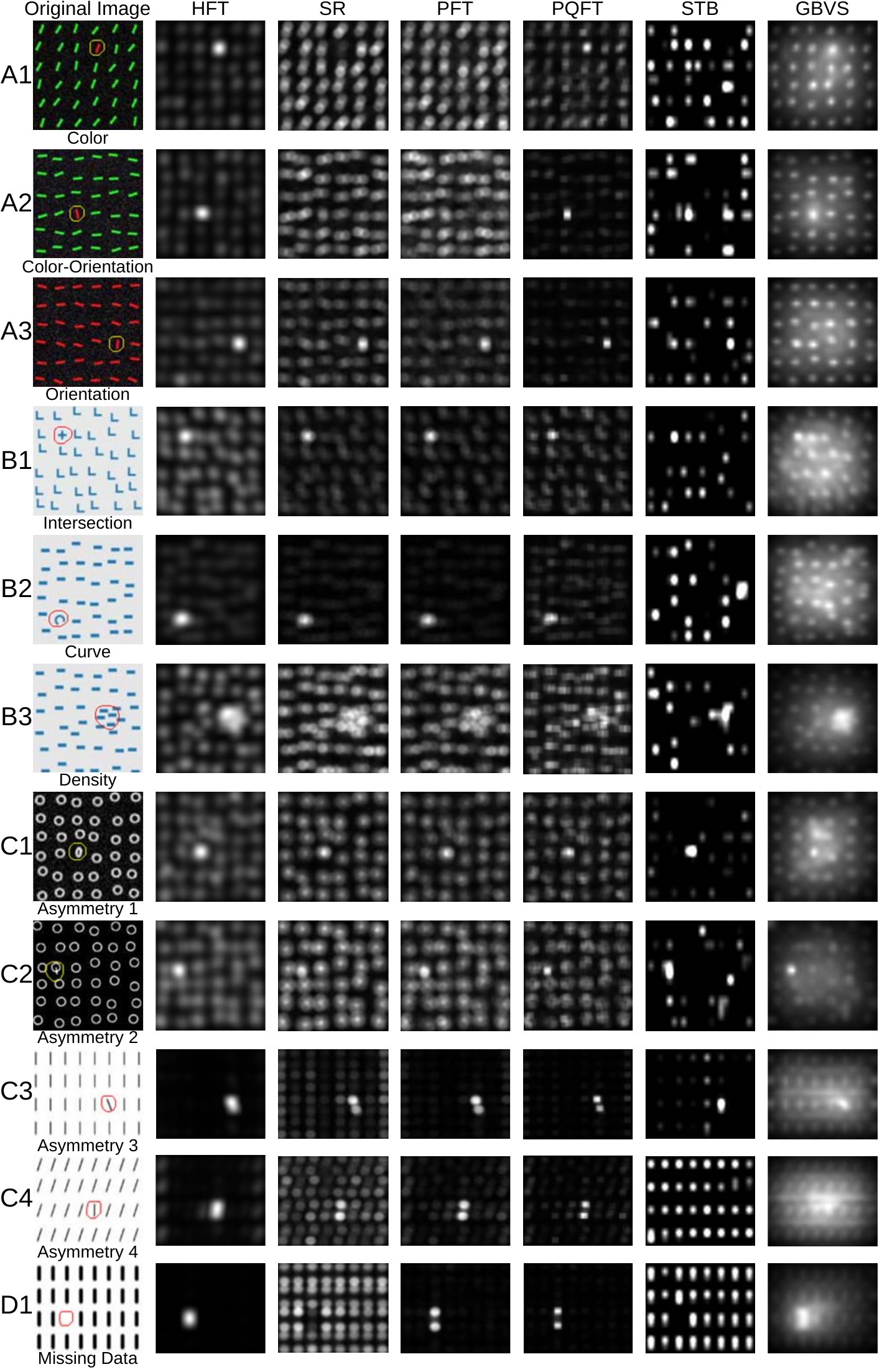}
\end{overpic}
\caption{Responses to so-called psychological patterns. The first column shows the boundary of the primary object computed by HFT, superimposed on the original image. For comparison, the remaining columns present the results obtained by the methods mentioned earlier.}
\label{fig:Exp1}
\end{center}
\end{figure}

Four kinds of psychological patterns are employed: salient orientation and colored patterns (part A  in Fig. \ref{fig:Exp1}), salient shape patterns (part B), asymmetric patterns (part C) and patterns with missing items (part D). The first column in Fig. \ref{fig:Exp1} shows the original images and the second shows the saliency maps produced by HFT. The proto-objects given by HFT are superimposed on the original images in the first column. Our results are compared with SR, PFT, PQFT, STB and GBVS.

As shown in part A of Fig. \ref{fig:Exp1}, the first pattern is salient due to the distinguishing color. Both HFT and PQFT obtain acceptable results, while SR, PFT and STB are unable to highlight the salient bar. The second image contains a salient bar having both different color and orientation. HFT, GBVS and PQFT succeed in highlighting the red bar, while the other methods fail. The salient bar in the third image has a distinguishing orientation, and only GBVS failed to locate it.

In part B of Fig. \ref{fig:Exp1}, HFT, SR, PFT and PQFT function well. However, in B3, although both SR and PFT are able to highlight the salient region, the so-called common regions are not suppressed correctly. In B2, STB highlights the wrong area and in B3, both STB and PQFT cannot detect the salient region.

All of the algorithms are able to find the asymmetric salient regions in C1-C2. However, the common regions are not suppressed sufficiently by the SR, PFT, GBVS and PQFT. All of the algorithms perform well for C3, although HFT achieves the best result. Finding the salient bar in C4 is apparently a more difficult task for humans and this also seems to be the case for SR, PFT and PQFT. In general, the results are not as good as those for C3, and STB and GBVS has even failed completely.

Sometimes a salient region is simply an empty area, as shown in D1 of Fig. \ref{fig:Exp1}. A good salient detector should be able to locate such a region as well. We find that HFT, PFT, GBVS and PQFT can detect the missing item successfully, although HFT does a better job.

Overall HFT performs the best for all the cases shown in Fig. \ref{fig:Exp1}. We also note that SR and PFT  obtain more or less the same results in cases A1-C2. However, they produce different results in cases C3-D1\footnote{In section \ref{relatedworks}, we draw the conclusion that both SR and PFT will yield nearly the same results based on the assumption of a natural image input. However, for "unnatural" images, they will sometimes produce different results, as is also discussed in \cite{ guo2010multiresolution}.}. Since PQFT is an advanced version of PFT, its performance is an improvement over the latter, especially in the case of colored tokens. However, in the rest of the cases, PQFT achieves nearly the same performance as PFT and SR.

\begin{figure}[h]
\begin{center}
\begin{overpic}[width=6cm]{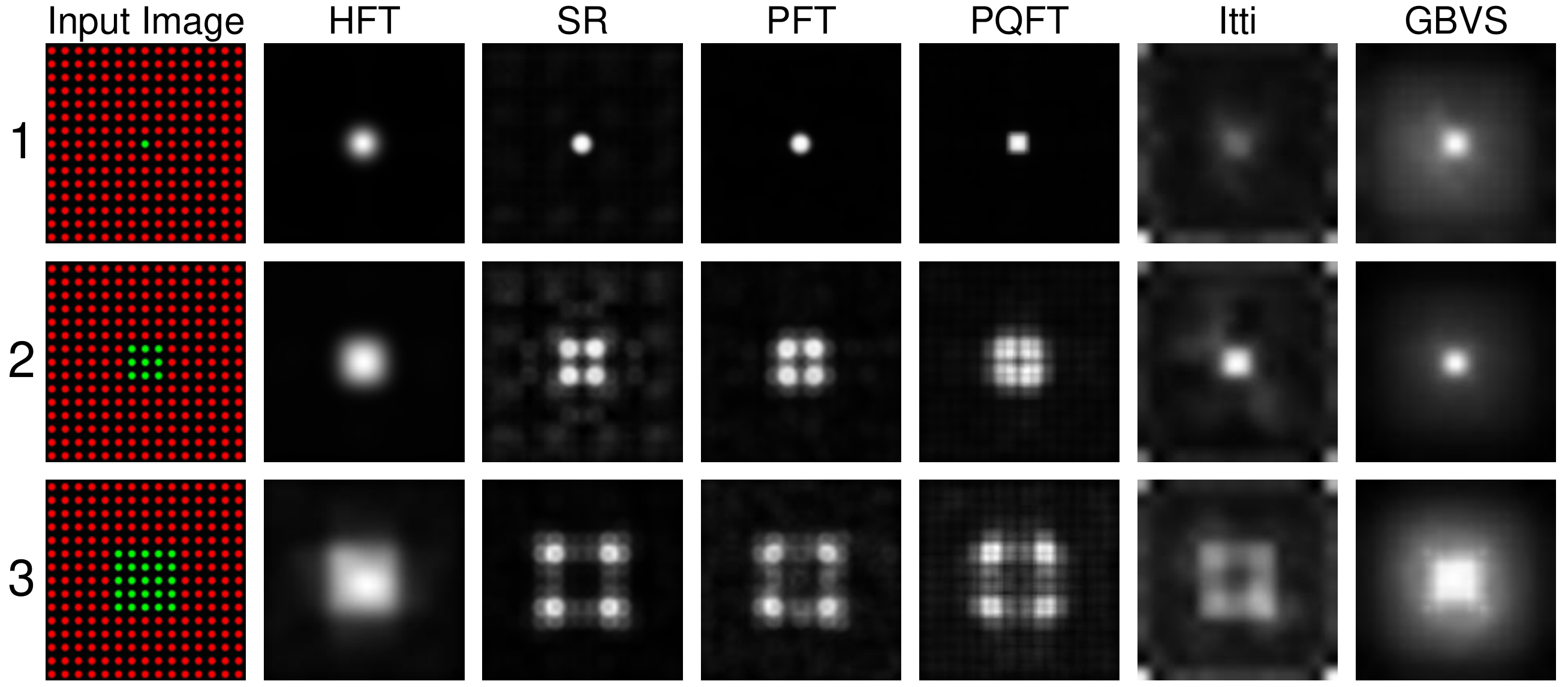}

\end{overpic}
\caption{Responses to psychological patterns with salient regions of different size.}
\label{fig:Exp2}
\end{center}
\end{figure}

We expect that a saliency detector would highlight salient regions of difference sizes in an image that people would pay attention to \cite{einhauser2008objects,LCAV-CONF-2009-012}. In order to examine this issue, we created three patterns in which the size of the tokens increased progressively, as shown in Fig. \ref{fig:Exp2}. All of the algorithms responded well to the small regions (see row 1). However, as the size increased, the performance of PQFT, Itti and SR/PFT are decreased. We observe that both SR/PFT and PQFT only respond to the boundaries of the regions when the salient region is large, while both HFT and GBVS highlight the salient region uniformly.

\begin{figure}[h]
\begin{center}
\begin{overpic}[width=6cm]{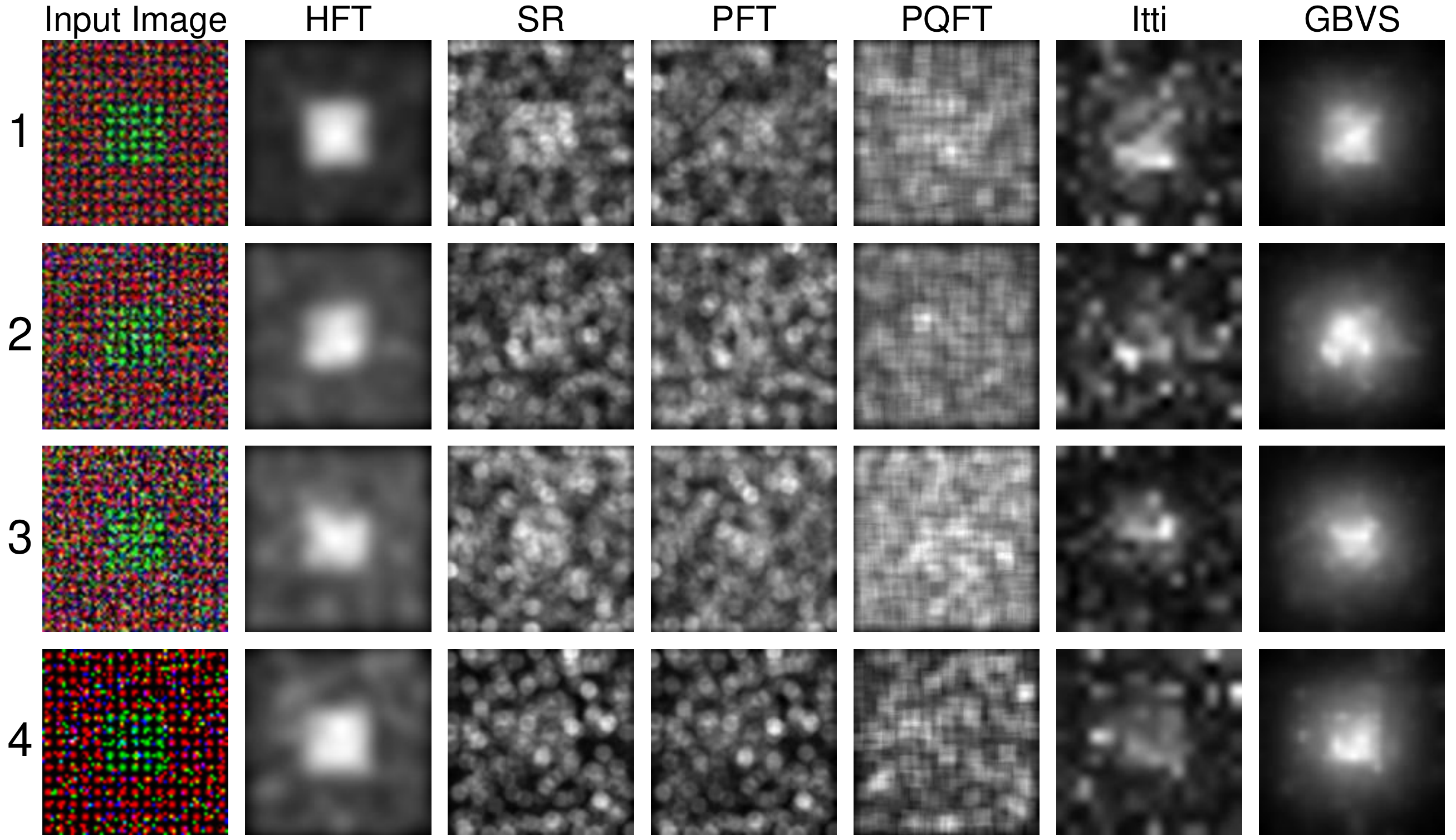}

\end{overpic}
\caption{Responses to psychological patterns with noise.}
\label{fig:Exp3}
\end{center}
\end{figure}

Finally, in order to evaluate the noise tolerance of each model, we added different amounts of  Gaussian (row 1-3 of  Fig. \ref{fig:Exp3}) and salt$\&$pepper (row 4) noise to the pattern. As shown in Fig. \ref{fig:Exp3}, the proposed HFT obtained the best overall performance, while GBVS also performed quite well. GBVS is an improved version of Itti's saliency model, and its anti-noise properties have also improved. We observe that SR/PFT and PQFT are quite sensitive to both Gaussian and salt and pepper noise.

\subsection{Predicting Human Attention Using Fixations}
\label{exp_fixation}

\begin{figure}[h]
\begin{center}
\begin{overpic}[width=7.5cm]{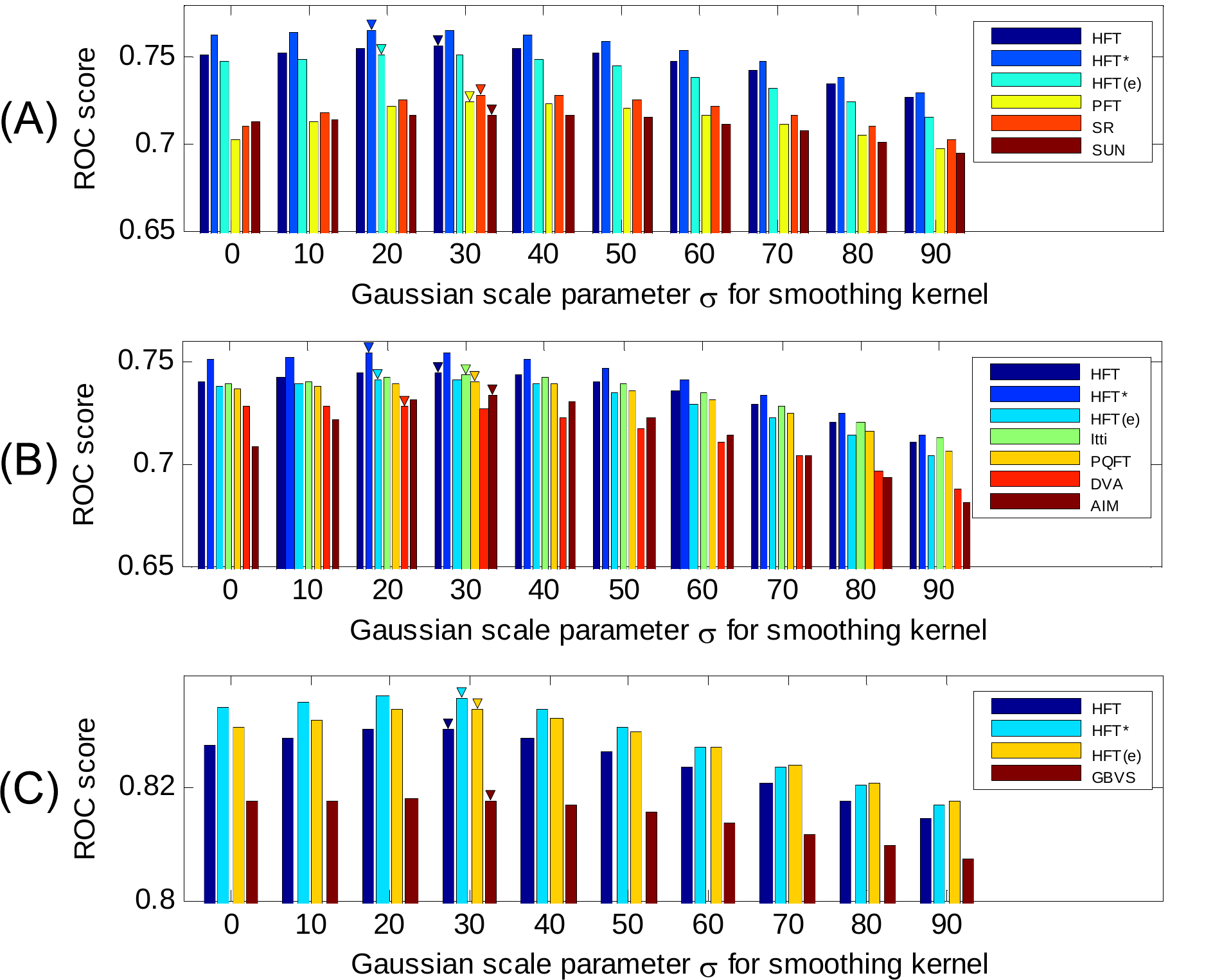}

\end{overpic}
\caption{Performance (peak ROC score) comparison between the HFT class models and those in the three subsets. (A)Comparing HFT with models in subset 1; (B)Comparing HFT with models in subset 2; (B)Comparing HFT with models in subset 3.} 
\label{fig:Exp33}
\end{center}
\end{figure}

We have evaluated HFT and compared it with state-of-the-art methods using human fixation data. Bruce's database\cite{NIPS2005_81} was employed for this purpose. It includes 120 natural images as well as corresponding eye-tracking data. The {\it quantitative} results are shown in Fig. \ref{fig:Exp33}.

We first compare HFT class models with models in subset 1. There is no border cut and center-bias in these models, so we need only find the optimal smoothing scale to compare the models. Fig. \ref{fig:Exp33}(A) shows the ROC scores for each model with different smoothing scales; we observe that they achieve their maximum ROC scores at different smoothing levels. We use the peak ROC score to establish the performance of each model, so that the influence of smoothing is compensated. In Fig.\ref{fig:Exp33}(A)-(C), the peak performance of each algorithm is labeled by a triangle. As shown in Fig. \ref{fig:Exp33}(A), it is obvious that HFT  obtains the best performance, while SR, PFT and SUN have about the same performance.


When comparing HFT with models in subset 2, the border cut is set for both HFT class models and the models in the subset. In fact, none of the models employ the same border cut, and therefore are not immediately comparable. Without calibration, the models in subset 2 will have a mendacious ROC score. As shown in Fig. \ref{fig:Exp33}(B), both HFT class algorithms and Itti's model are the highest performing models, while AIM has a higher peak performance than DVA.

In the more recent literature, GBVS always yields a very high ROC score and outperforms other models. However,this is most likely because GBVS incorporates a global center-bias \cite{judd2012pami}. When comparing HFT class models with GBVS, we selected the optimal center-bias for both. We note from Fig.\ref{fig:Exp33}(C) that HFT class algorithms achieve quite a high performance level, all of them outperforming GBVS.

\subsection{Predicting Salient Regions That Humans Attend}
\label{exp_region}
Besides using fixation data, we also used object maps labeled by humans to evaluate the algorithms (Refer to \url{http://www.cim.mcgill.ca/~lijian} for details of how this database was obtained).
Obviously, different images present different levels of difficulty for any saliency detector. However, the existing saliency benchmarks in the literature are collections of images, with no attempt to categorize the difficulty of analysis required. In this paper, a database containing 235 images was collected using Google as well as the recent literature. The images in this database were divided into 6 categories: 1) 50 images with large salient regions; 2) 80 images with intermediate salient regions; 3) 60 images with small salient regions; 4) 15 images with cluttered backgrounds; 5) 15 images with repeating distractors; 6) 15 images with both large and small salient regions. In this section, we report both the overall performance of each model as well as the performance of each model for each category.

\begin{figure*}
\begin{center}
   \includegraphics[width=.97\linewidth]{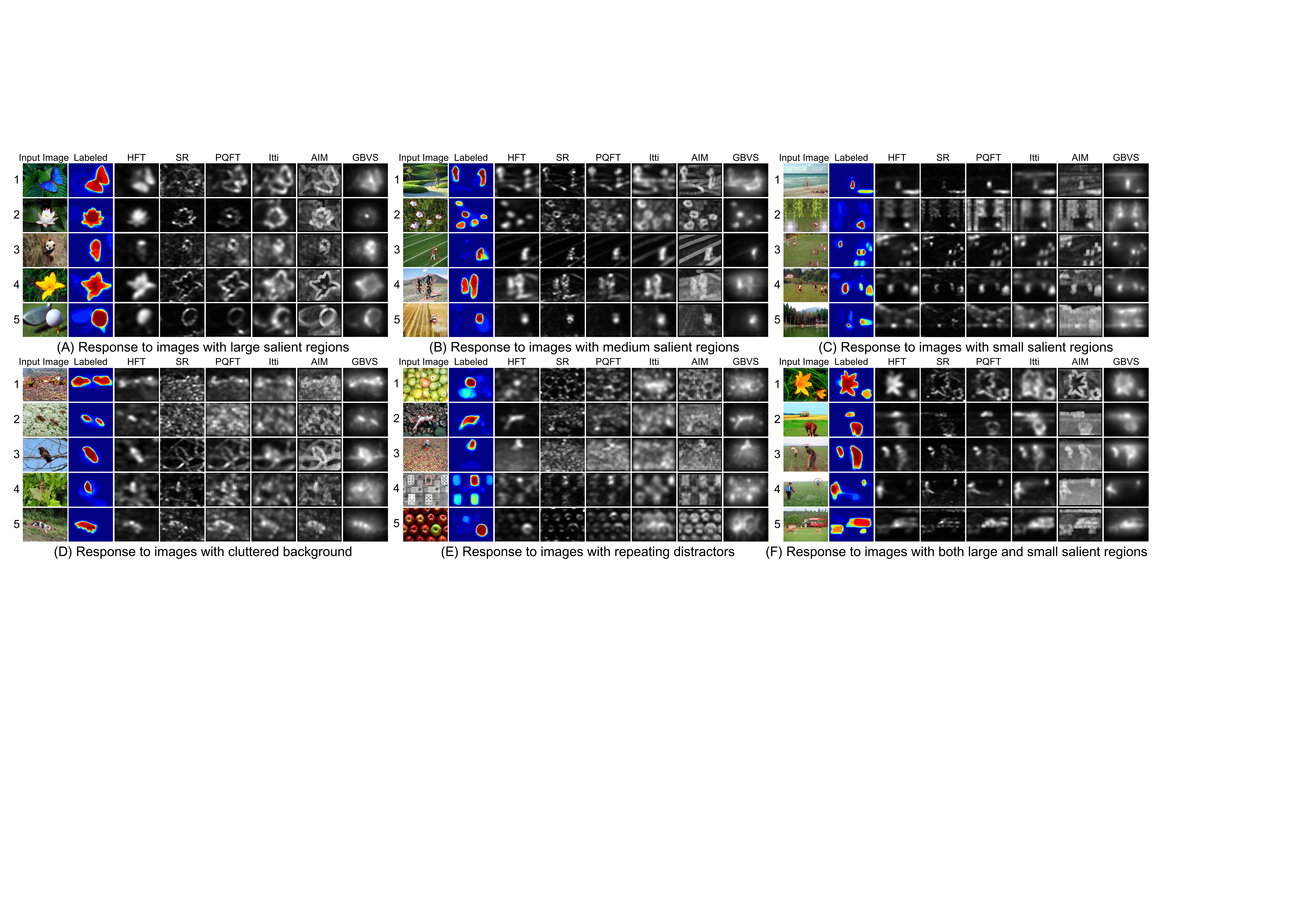}
\end{center}
   \caption{Procedure for computing  saliency  using the Hypercomplex Fourier Transform (HFT)}
\label{fig:Exp71}
\end{figure*}

In this experiment, we report only the performance at the optimal smoothing level. Both the ROC score (AUC) and the peak value of the DSC curve (PoDSC) for each model were calculated as shown in Table \ref{Smooth}-\ref{GBVS}. Fig.\ref{fig:Exp71} shows some examples which permit a qualitative comparison for each category in the dataset. However, due to space limitations, we are unable to show all of the {\it qualitative} results. However, these are available at \url{http://www.cim.mcgill.ca/~lijian}.

Fig.\ref{fig:Exp71}(A) shows natural images with large salient regions, a situation that is challenging for many models. It is clear that HFT achieves the best performance. The AUC and PoDSC criteria also support this conclusion. We note that GBVS achieves reasonable results, but SR, PQFT and AIM only enhance the boundaries instead of highlighting the whole salient region uniformly.

In Fig.\ref{fig:Exp71}(B), there are five images with intermediate salient regions. For example, in the second one, there are five salient flowers in the scene. HFT and GBVS have detected these object regions correctly. However, all of the other models failed to highlight them uniformly.

The images in  Fig.\ref{fig:Exp71}(C) contain distant objects and distractors (e.g., the skyline in row 1). We observe that most of the algorithms work well in detecting the small salient regions. However, sometimes Itti's method and GBVS fail to suppress distractors (see row 1). HFT only slightly outperforms the others for this category.

The backgrounds of the images in Fig.\ref{fig:Exp71}(D) are quite cluttered. This case is also difficult because many algorithms are quite sensitive to background noise. For example, consider the image in row 2. For SR, PQFT, Itti and AIM, the non-salient grassy surface of the ground is enhanced as much as the two insects. However, HFT and GBVS detect these two regions correctly. HFT achieves excellent performance in this category, which is also supported by the quantitative results.

Compared to salient objects, repeating distractors in the scene should not attract much attention from humans \cite{beck2005stimulus}. Fig.\ref{fig:Exp71}(E) shows images with salient objects among repeating objects. In both rows 2 and 3, HFT and GBVS suppress the repeating objects and enhance the salient object correctly. In row 4, there is a salient playing card among the five, and HFT, PQFT and SR highlight the salient one and suppress the other four. HFT achieved the best performance for this category as well.

If an image contains both large and small salient regions (see Fig.\ref{fig:Exp71}(F)), a detector should be able to detect both simultaneously. For example, in row 1, there are two flowers of different size but SR, PQFT, Itti and AIM only respond strongly on their boundaries. However, HFT and GBVS respond correctly. Nevertheless, as discussed earlier, HFT selects just one optimal scale to determine the final output. Hence, objects of different size are not all detected or enhanced uniformly in the saliency map (see row 5 of Fig.\ref{fig:Exp71}(F)).

Overall, the experimental results shown in Table \ref{Smooth}-\ref{GBVS} indicate that the HFT model achieves the best performance for all six categories. Moreover, HFT exhibits superior performance when detecting large salient regions and saliency in cluttered scenes.

As mentioned earlier, images in this database contain salient regions of different sizes. Interestingly, in \cite{hou2007saliency} it is suggested that in order to find objects at different scales it should be possible to use different resolutions of the input image. In order to investigate this issue, we created different resolutions of the input image and then fed them into SR and the other one-resolution models (see Table \ref{one-scale}). We used the criterion described in (\ref{Eq:pscal5}) to find the optimal saliency map as the final output. We note that the performance of these revised models has improved (as shown in Table \ref{one-scale}), although it is still lower than the HFT class models (See Table \ref{Smooth}-\ref{GBVS}.)

\begin{table}
\renewcommand{\arraystretch}{1.3}
\caption{Performance of the revised one-resolution models}
\label{one-scale}
\centering
\begin{tabular}{lcc}
\toprule
Model  & \bfseries AUC (improvement) & \bfseries PoDSC (improvement)\\
\hline
SR   & 0.8733 ($\uparrow$ 0.0210) & 0.4316 ($\uparrow$ 0.0387)\\
PFT  & 0.8769 ($\uparrow$ 0.0243) & 0.4420 ($\uparrow$ 0.0456)\\
PQFT & 0.8951 ($\uparrow$ 0.0197) & 0.4963 ($\uparrow$ 0.0234)\\
SUN  & 0.8470 ($\uparrow$ 0.0067) & 0.4139 ($\uparrow$ 0.0298)\\
AIM  & 0.8858 ($\uparrow$ 0.0027) & 0.4992 ($\uparrow$ 0.0050)\\
\bottomrule
\end{tabular}
\end{table}

\begin{figure}
\begin{center}
   \includegraphics[width=.8\linewidth]{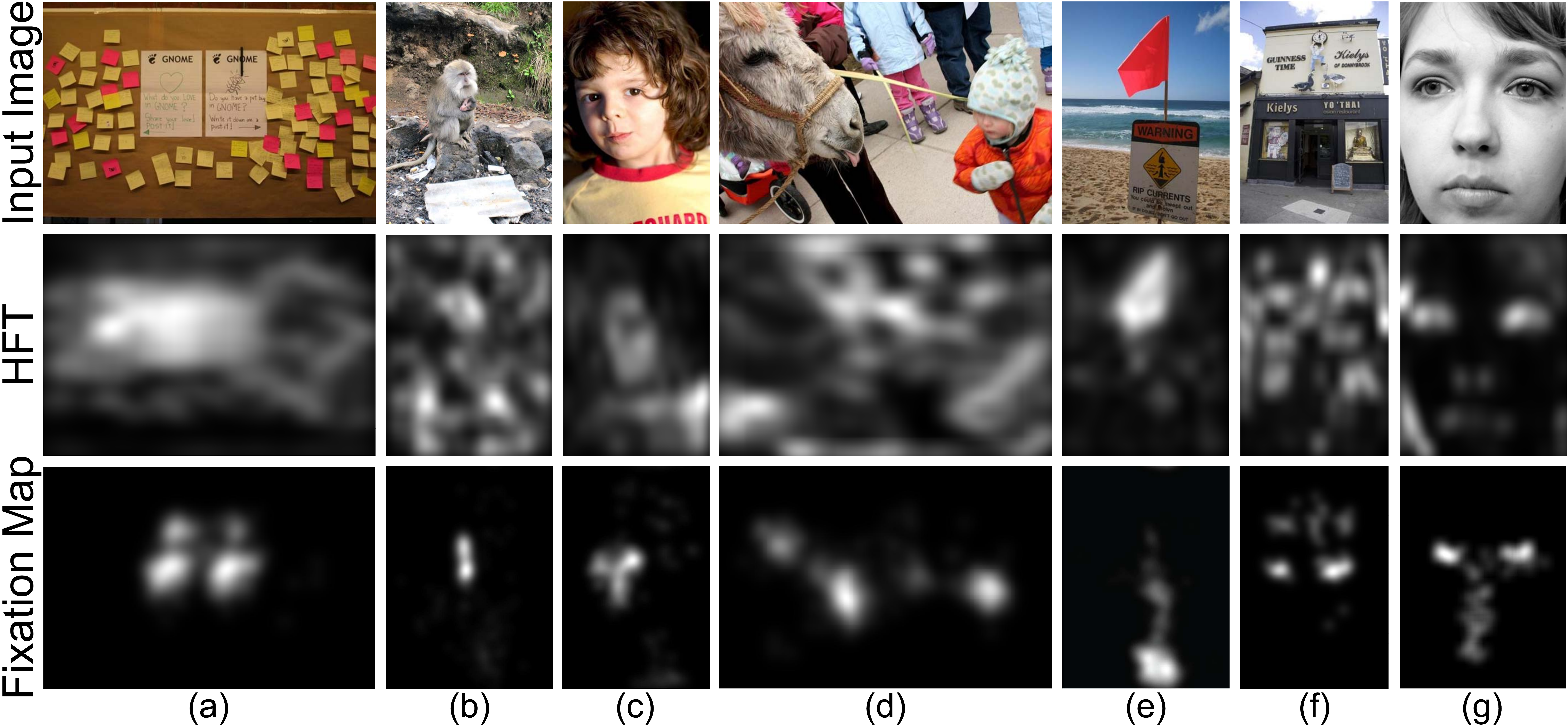}
\end{center}
   \caption{Hard image cases of HFT in predicting human fixations}
\label{failcases}
\end{figure}

Although HFT has performed well in the  experiments described in section \ref{EXP:Psy}
 to \ref{exp_region}, it does fail in certain cases.   HFT could not satisfactorily predict the correct human fixations  for several of the "hard" images collected in \cite{judd2012pami}. HFT did predict the human fixations correctly in Fig.\ref{failcases} (a, b), although some incorrect responses did occur. However, HFT did a poor job for some images. For example, in (c) it incorrectly highlighted some parts of the clothes and failed to highlight the eyes, while in (g), some parts of the boundary  of the face were wrongly highlighted. In both (e) and (f), people tended to pay attention to the text, but HFT locates regions with salient low-level features (e.g., the red flag and the clock). In (d), HFT totally failed to locate the salient heads.
Clearly prior knowledge and task information is not employed for bottom-up models. Therefore, these approaches focus on regions possessing distinct low-level  features (color, intensity etc.) and sometimes may fail to highlight the regions that are known to interest people (e.g., humans, animals and other common objects). One way to solve this problem is to employ more complex features or invoke top-down cues.


Most of the bottom-up saliency models, such as Itti, Gao's model, AIM and so on, use local contrast or center surround paradigm. Similarly, models like SR can also be considered as pixel-level local contrast models (gradient operation). These work well for detecting small salient regions, but do not perform well in predicting large salient regions. There are two ways to alleviate this problem; one is to adopt a multi-scale strategy (as used in Itti's model), the other is to decrease the resolution of the input image and employ a large amount of blurring of the saliency maps (as used in SR, PFT).
Finally, perhaps it is unfair to characterize SR class models as being only pixel-level local contrast deterctors. As discussed earily, SR and PFT are special cases of the proposed HFT model when the scale goes to infinity in the frequency domain. Hence they have the ability to globally inhibit and suppress repeated patterns. However, other models based on local contrast will perform poorly in this case. Nevertheless, if there are no repeated patterns in the scene, the SR model will function as a gradient detector, and only enhance boundaries of objects.


\begin{table*}[htbp] 	
\caption{Comparison between HFT class models and models in subset 1 (optimal smoothing parameters for each algorithm)} 	
\label{Smooth}
\centering 	
\begin{tabular}{@{}lp{.78cm}p{.78cm}p{.78cm}p{.78cm}p{.78cm}p{.78cm}p{.78cm}p{.78cm}p{.78cm}p{.78cm}p{.78cm}p{.78cm}p{.78cm}p{.78cm}@{}}
\toprule
&\multicolumn{2}{c}{Category 1}
&\multicolumn{2}{c}{Category 2}
&\multicolumn{2}{c}{Category 3}
&\multicolumn{2}{c}{Category 4}
&\multicolumn{2}{c}{Category 5}
&\multicolumn{2}{c}{Category 6}
&\multicolumn{2}{c}{Overall}\\
\cmidrule(lr){2-3} \cmidrule(lr){4-5} \cmidrule(lr){6-7} \cmidrule(lr){8-9} \cmidrule(lr){10-11} \cmidrule(lr){12-13} \cmidrule(lr){14-15}
Model & \scriptsize{AUC} & \scriptsize{PoDSC}
& \scriptsize{AUC} & \scriptsize{PoDSC}
& \scriptsize{AUC} & \scriptsize{PoDSC}
& \scriptsize{AUC} & \scriptsize{PoDSC}
& \scriptsize{AUC} & \scriptsize{PoDSC}
& \scriptsize{AUC} & \scriptsize{PoDSC}
& \scriptsize{AUC} & \scriptsize{PoDSC}\\
\midrule 	
%

{\bf HFT} &{\bf0.9424} &{\bf0.7252}        &{\bf0.9146} &{\bf0.5481}        &{\bf0.9351} &{\bf0.4563}        &{\bf0.9448} &{\bf0.5856}        &{\bf0.9193} &{\bf0.5778}        &{\bf0.9535} &{\bf0.6976}  &{\bf 0.9281} &{\bf 0.5417}\\
HFT(e)&0.9101 &0.6592        &0.9050 &0.5112        &0.9348 &0.4502        &0.9463 &0.6306        &0.8907 &0.5123        &0.9418 &0.6489  &0.9159 &0.5029\\
HFT* &0.9543 &0.7438        &0.9425 &0.6184        &0.9572 &0.5217        &0.9686 &0.6846        &0.9413 &0.6148        &0.9709 &0.7568  &0.9497  &0.5963\\
SR	 &0.8148 &0.5104         &0.8495 &0.4321         &0.9091 &0.3281         &0.7595 &0.2796         &0.7929 &0.3266         &0.8924 &0.5404  &0.8523 &0.3929\\ 	
PFT	 &0.8064 & 0.5029 & 0.8426 & 0.4292 & 0.9269 & 0.3780 & 0.7294 & 0.2931 & 0.7724 & 0.3377 & 0.8967 & 0.5574 &0.8526 &0.3964\\ 	
SUN	 &0.8218 &0.5393       &0.8457 &0.4522        &0.8838 &0.3026        &0.6994 &0.2452        &0.8067 &0.3773        &0.8778 &0.5555 &0.8403  &0.4018\\
\bottomrule 	
\end{tabular} 	
\end{table*}

\begin{table*}[htbp] 	
\caption{Comparison between HFT class models and models in subset 2 (optimal smoothing parameters and the same border cut for each model)}
\label{AIM}	
\centering 	
\begin{tabular}{@{}lp{.78cm}p{.78cm}p{.78cm}p{.78cm}p{.78cm}p{.78cm}p{.78cm}p{.78cm}p{.78cm}p{.78cm}p{.78cm}p{.78cm}p{.78cm}p{.78cm}@{}}
\toprule
&\multicolumn{2}{c}{Category 1}
&\multicolumn{2}{c}{Category 2}
&\multicolumn{2}{c}{Category 3}
&\multicolumn{2}{c}{Category 4}
&\multicolumn{2}{c}{Category 5}
&\multicolumn{2}{c}{Category 6}
&\multicolumn{2}{c}{Overall}\\
\cmidrule(lr){2-3} \cmidrule(lr){4-5} \cmidrule(lr){6-7} \cmidrule(lr){8-9} \cmidrule(lr){10-11} \cmidrule(lr){12-13} \cmidrule(lr){14-15}
Model & \scriptsize{AUC} & \scriptsize{PoDSC}
& \scriptsize{AUC} & \scriptsize{PoDSC}
& \scriptsize{AUC} & \scriptsize{PoDSC}
& \scriptsize{AUC} & \scriptsize{PoDSC}
& \scriptsize{AUC} & \scriptsize{PoDSC}
& \scriptsize{AUC} & \scriptsize{PoDSC}
& \scriptsize{AUC} & \scriptsize{PoDSC}\\
\midrule

{\bf HFT} &{\bf0.9338}   &{\bf0.7395}   &{\bf0.9064}   &{\bf0.5697}   &{\bf0.9328}   &{\bf0.4871}  &{\bf0.9378}   &{\bf0.6074}     &{\bf0.9137}    &{\bf0.5893}   &{\bf0.9441}   &{\bf0.7114}   &{\bf0.9217}  &{\bf0.5627}\\
HFT(e)   &0.9010   &0.6867   &0.9004   &0.5402    &0.9312   &0.4621   &0.9471   &0.6568   &0.8660  &0.5286  &0.9340   &0.6743  &0.9102    &0.5289\\
HFT*   &0.9478  &0.7584    &0.9387  &0.6434    &0.9578   &0.5503   &0.9660   &0.7012   &0.9374    &0.6330    &0.9644   &0.7658   &0.9470  &0.6226\\
AIM &0.8511 &0.6011   &0.8761 &0.5226   &0.9359 &0.4506   &0.8370 &0.3969  &0.8668 &0.4987 &0.9124 &0.6489  &0.8831  &0.4942\\
PQFT &0.8571 &0.6201 &0.8771 &0.5350 &0.9096 &0.3901 &0.8205 &0.3819 &0.8421 &0.4304 &0.9105 &0.6398 &0.8754 &0.4729\\
DVA  &0.8075 &0.5736   &0.8565 &0.5095   &0.9038 &0.3957   &0.7618 &0.3639   &0.8250 &0.4553   &0.9048 &0.6262   &0.8510   &0.4642\\
Itti &0.8768 &0.6533 &0.8886 &0.5317 &0.9239 &0.3843 &0.8107 &0.3687 &0.8983 &0.5194 &0.9191 &0.6530 &0.8910 &0.4949\\
\bottomrule 	
\end{tabular} 	
\end{table*}

\begin{table*}[htbp] 	
\caption{Comparison between HFT class models and model in subset 3 (optimal smoothing parameters and center-bias for each model)}
\label{GBVS}	
\centering 	
\begin{tabular}{@{}lp{.78cm}p{.78cm}p{.78cm}p{.78cm}p{.78cm}p{.78cm}p{.78cm}p{.78cm}p{.78cm}p{.78cm}p{.78cm}p{.78cm}p{.78cm}p{.78cm}@{}}
\toprule
&\multicolumn{2}{c}{Category 1}
&\multicolumn{2}{c}{Category 2}
&\multicolumn{2}{c}{Category 3}
&\multicolumn{2}{c}{Category 4}
&\multicolumn{2}{c}{Category 5}
&\multicolumn{2}{c}{Category 6}
&\multicolumn{2}{c}{Overall}\\
\cmidrule(lr){2-3} \cmidrule(lr){4-5} \cmidrule(lr){6-7} \cmidrule(lr){8-9} \cmidrule(lr){10-11} \cmidrule(lr){12-13} \cmidrule(lr){14-15}
Model & \scriptsize{AUC} & \scriptsize{PoDSC}
& \scriptsize{AUC} & \scriptsize{PoDSC}
& \scriptsize{AUC} & \scriptsize{PoDSC}
& \scriptsize{AUC} & \scriptsize{PoDSC}
& \scriptsize{AUC} & \scriptsize{PoDSC}
& \scriptsize{AUC} & \scriptsize{PoDSC}
& \scriptsize{AUC} & \scriptsize{PoDSC}\\
\midrule
{\bf HFT}       &{\bf0.9565} &{\bf0.7548} &{\bf0.9296} &{\bf0.5688} &{\bf0.9504} &{\bf0.4755} &{\bf0.9381} &{\bf0.5799} &{\bf0.9523} &{\bf0.6318} &{\bf0.9578} &{\bf0.7020} &{\bf0.9414} &{\bf0.5665}\\
HFT(e)    &0.9409 &0.7213 &0.9287 &0.5697 &0.9614 &0.5024 &0.9460 &0.6315 &0.9361 &0.6038 &0.9579 &0.7089 &0.9403 &0.5609\\
HFT*      &0.9665 &0.7771 &0.9533 &0.6396 &0.9724 &0.5640 &0.9709 &0.7028 &0.9644 &0.6841 &0.9743 &0.7586 &0.9609 &0.6250\\
GBVS      &0.9363 &0.6990 &0.9135 &0.5304 &0.9173 &0.3678 &0.9223 &0.5644 &0.9453 &0.6145 &0.9249 &0.6329 &0.9211 &0.5154\\
\bottomrule 	
\end{tabular} 	
\end{table*}

\section{Conclusions}

This paper proposes a new saliency detection framework for images, based on analyzing the spectrum scale-space. We show that the convolution of the image amplitude spectrum with a low-pass Gaussian kernel of an appropriate scale is equivalent to such an image saliency detector. The proposed approach is able to highlight both small and large salient regions and inhibit repeated patterns. We also illustrate that both SR and PFT are special cases of the proposed model when the scale parameter goes to infinity.
In order to fuse multi-dimensional feature maps, we employ the Hypercomplex Fourier Transform to replace the standard Fourier Transform for spectrum scale-space analysis.

To validate the proposed approach, we have performed saliency computations on both commonly used synthetic data as well as natural images, and then compared the results with state-of-the-art algorithms. In order to make a fair comparison when using the ROC and PoDSC as measures of performance, we have proposed an improved comparison procedure by considering the border cut, center-bias and smoothing effects. Experimental results indicate that the proposed model can predict human fixation data as well as the object regions labeled by humans. We also show that sometimes HFT may fail to predict human fixations. This is most likely because only low-level features are employed; Clearly, top-down or task-orientated cues are necessary for improving the performance of current saliency models in predicting human attention.

With regard to future work, firstly, it would be interesting to investigate other criteria for optimal scale selection. Entropy was employed in this paper to select the optimal scale automatically. However, we have observed that a much higher performance can be obtained by selecting the optimal scale manually. Secondly, in the proposed model, only one saliency map, corresponding to the optimal scale, is selected as the final one. However, we have noted that certain of the abandoned maps also contain meaningful saliency information. How to incorporate these in the determination of the saliency is left to future investigation. Thirdly, we intend to include top-down information to improve performance. The ultimate goal of our research is to develop a system for on-board pedestrian and vehicle detection, for which a considerable amount of top-down temporal data exists.
{\small
\bibliographystyle{IEEEtran}
\bibliography{mybib}
}
\begin{IEEEbiography}[{\includegraphics[width=1in,height=1.25in,clip,keepaspectratio]{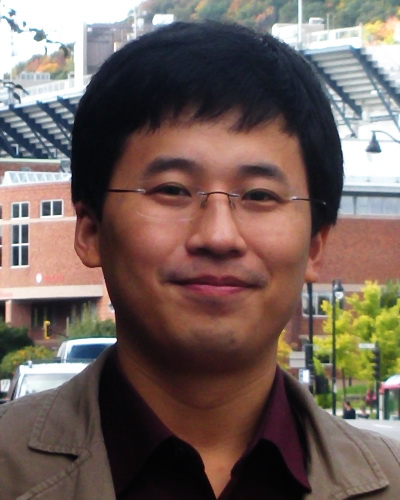}}]{Jian Li}
received the B.E. degree and the M.E. degree in control science and engineering from National University of Defense Technology (NUDT), Changsha, Hunan, P.R. China, where he is currently working toward the Ph.D. degree. He is also a visiting Ph.D. student at Center for Intelligent Machines (CIM) in McGill University. 	

His research interests include computer vision, pattern recognition, image processing, and machine learning.
\end{IEEEbiography}
\begin{IEEEbiography}[{\includegraphics[width=1in,height=1.25in,clip,keepaspectratio]{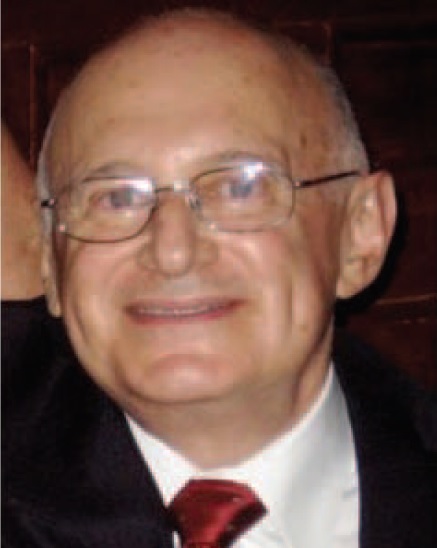}}]{{Martin D. Levine}}
received
the B. Eng. and M. Eng. degrees in Electrical and Computer Engineering from McGill University, Montreal, in 1960 and 1963, respectively, and the Ph.D. degree in Electrical Engineering from the Imperial College of Science and Technology, University of London, London, England, in 1965. He is currently a Professor in the Department of Electrical and Computer Engineering, McGill University and served as the founding Director of the McGill Center for Intelligent Machines (CIM) from 1986 to 1998. During 1972-1973 he was a member of the Technical Staff at the Image Processing Laboratory of the Jet Propulsion Laboratory, Pasadena, CA. During the 1979-1980 academic year, he was a Visiting Professor in the Department of Computer Science, Hebrew University, Jerusalem, Israel. His research interests include computer vision, image processing and artificial intelligence, and he has numerous publications to his credit on these topics. As well, he has consulted for various government agencies and industrial organizations in these areas. Dr. Levine was a founding partner of AutoVu Technologies Inc. and Vision-Sphere Technologies Inc. Dr. Levine has authored the book entitled Vision in Man and Machine and has co-authored Computer Assisted Analyses of Cell Locomotion and Chemotaxis. Dr. Levine is an Area Editor for face detection and recognition and on the Editorial Board of the journal Computer Vision and Understanding, having also served on the Editorial Boards of the IEEE TRANSACTIONS ON PATTERN ANALYSIS AND MACHINE INTELLIGENCE and Pattern Recognition. He was the Editor of the Plenum Book Series on Advances in Computer Vision and Machine Intelligence. He was the General Chairman of the Seventh International Conference on Pattern Recognition held in Montreal during the summer of 1984 and served as President of the International Association of Pattern Recognition during 1988-1990. He was also the founding President of the Canadian Image Processing and Pattern Recognition Society. Dr. Levine was elected as a Fellow of the Canadian Institute for Advanced Research in 1984. During the period 1990-96 he served as a CIAR/PRECARN Associate. He is a Fellow of the IEEE, the Canadian Academy of Engineering and the International Association for Pattern Recognition. Dr. Levine was presented with the 1997 Canadian Image Processing and Pattern Recognition Society Service Award for his outstanding contributions to research and education in Computer Vision.
\end{IEEEbiography}
\begin{IEEEbiography}[{\includegraphics[width=1in,height=1.25in,clip,keepaspectratio]{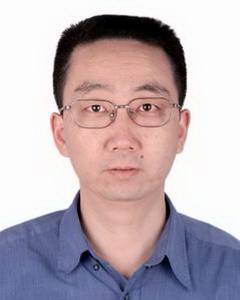}}]{Xiangjing An}
received the B.S. degree in automatic control from the Department of Automatic Control, National University of Defense Technology (NUDT), Changsha, P. R. China, in 1995 and the Ph.D. degree in control science and engineering from the College of Mechatronics and Automation (CMA), NUDT in 2001. He has been an visiting scholar for cooperation research in the Boston University during 2009-2010. Currently, he is an Associate Professor at the Institute of Automation, CMA, NUDT. His research interests include image processing, computer vision, machine learning and biologically inspired feature extraction.
\end{IEEEbiography}
\begin{IEEEbiography}[{\includegraphics[width=1in,height=1.25in,clip,keepaspectratio]{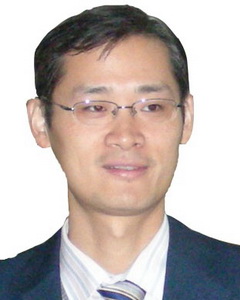}}]{Xin Xu}
received the B.S. degree in electrical engineering from the Department of Automatic Control, National University of Defense Technology (NUDT), Changsha, P. R. China, in 1996 and the Ph.D. degree in control science and engineering from the College of Mechatronics and Automation (CMA), NUDT. He has been a visiting scientist for cooperation research in the Hong Kong Polytechnic University, University of Alberta, University of Guelph, and the University of Strathclyde, respectively. Currently, he is a Full Professor at the Institute of Automation, CMA, NUDT. He has coauthored four books and published more than 70 papers in international journals and conferences. His research interests include reinforcement learning, learning control, robotics, data mining, autonomic computing, and computer security.
Dr. Xu is one of the recipients received the 1st class Natural Science Award from Hunan Province, P. R. China, in 2009 and the Fork Ying Tong Youth Teacher Fund of China in 2008. He is a Committee Member of the IEEE Technical Committee on Approximate Dynamic Programming and Reinforcement Learning (ADPRL) and the IEEE Technical Committee on Robot Learning. He has served as a PC member or Session Chair in many international conferences.
\end{IEEEbiography}
\begin{IEEEbiography}[{\includegraphics[width=1in,height=1.25in,clip,keepaspectratio]{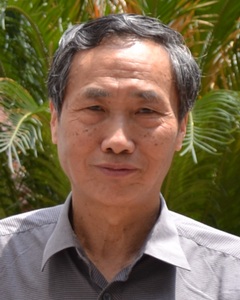}}]{Hangen He}
received the BSc degree in Nuclear Physics from Harbin Engineering Institute, Harbin, China, in 1968. He was a visiting Professor at the University of the German Federal Armed Forces in 1996 and 1999 respectively. He is currently a professor in the College of Mechatronics and Automation (CMA), National University of Defense Technology (NUDT), Changsha, Hunan, China. His research interests include artificial intelligence, reinforcement learning, learning control and robotics. He has served as a member of editorial boards of several journals and has cochaired many professional conferences. He is a joint recipient of more than a dozen academic awards in China.
\end{IEEEbiography}
\end{document}